\documentclass{article}

\usepackage[final, nonatbib]{neurips_2021}

\usepackage[utf8]{inputenc} 
\usepackage[T1]{fontenc}    
\usepackage[hidelinks]{hyperref} 
\usepackage{url}            
\usepackage{booktabs}       
\usepackage{amsfonts}       
\usepackage{nicefrac}       
\usepackage{microtype}      
\usepackage{xcolor}         
\usepackage{amsmath}
\usepackage{amsthm}
\usepackage{amssymb}
\usepackage{graphicx}
\usepackage{caption}
\usepackage{subcaption}
\graphicspath{{./figures/}}
\usepackage{etoolbox}

\makeatletter

\newcommand{\define}[4][ignore]{%
  \ifstrequal{#1}{ignore}{}{
  \@namedef{thmtitle@#2}{#1}}%
  \@namedef{thm@#2}{#4}%
  \@namedef{thmtypen@#2}{lemma}%
  \newtheorem{thmtype@#2}[theorem]{#3}%
  \newtheorem*{thmtypealt@#2}{#3~\ref{#2}}%
}

\newcommand{\state}[1]{%
  \@namedef{curthm}{#1}
  \@ifundefined{thmtitle@#1}{
  \begin{thmtype@#1}
    }{
  \begin{thmtype@#1}[\@nameuse{thmtitle@#1}]
  }
    \label{#1}
    \@nameuse{thm@#1}
  \end{thmtype@#1}
  \@ifundefined{thmdone@#1}{
  \@namedef{thmdone@#1}{stated}%
  }{}
}

\newcommand{\restate}[1]{%
  \@namedef{curthm}{#1}
  \@ifundefined{thmtitle@#1}{
    \begin{thmtypealt@#1}
    }{
  \begin{thmtypealt@#1}[\@nameuse{thmtitle@#1}]
  }
    \@nameuse{thm@#1}
  \end{thmtypealt@#1}
  \@ifundefined{thmdone@#1}{
  \@namedef{thmdone@#1}{stated}%
  }{}
}

\newcommand{\thmlabel}[1]{
  \@ifundefined{thmdone@\@nameuse{curthm}}{\label{#1}
    }{\tag*{\eqref{#1}}}
}
\makeatother

\usepackage{enumitem}
\usepackage{bm}

\newtheorem{theorem}{Theorem}[section]

\newtheorem{definition}[theorem]{Definition}

\newcounter{example}[section]

\newcommand{\norm}[1]{\|#1\|}

\newcommand{\wh}{\widehat}

\newcommand{\eps}{\varepsilon}

\newcommand{\R}{\mathbb{R}}
\newcommand{\C}{\mathbb{C}}

\newcommand{\nth}[1]{#1$^{\mathrm{th}}$}
\newcommand{\RN}[1]{%
  \textup{\uppercase\expandafter{\romannumeral#1}}%
}

\newcommand{\vertiii}[1]{{\left\vert\kern-0.25ex\left\vert\kern-0.25ex\left\vert #1 
		\right\vert\kern-0.25ex\right\vert\kern-0.25ex\right\vert}}

\DeclareMathOperator*{\E}{\mathbb{E}}

\DeclareMathOperator{\supp}{supp}
\DeclareMathOperator{\esssup}{ess\,sup}

\DeclareMathOperator*{\argmin}{arg\,min}

\newcommand{\xtilde}{\tilde{x}}

\newcommand{\xhat}{\wh{x}}
\newcommand{\zhat}{\wh{z}}

\newcommand{\cA}{\mathcal A}

\newcommand{\cM}{\mathcal M}
\newcommand{\cN}{\mathcal N}

\newcommand{\cW}{\mathcal W}

\DeclareMathOperator{\cov}{Cov}

\makeatletter
\newcommand{\printfnsymbol}[1]{%
  \textsuperscript{\@fnsymbol{#1}}%
}
\makeatother

\title{Robust Compressed Sensing MRI with Deep Generative Priors}

\author{%
Ajil Jalal\thanks{Ajil Jalal and Marius Arvinte contributed equally to this work.}\\
ECE, UT Austin \\ 
\texttt{ajiljalal@utexas.edu} 
\And
Marius Arvinte\printfnsymbol{1}  \\ 
ECE, UT Austin \\ 
\texttt{arvinte@utexas.edu}  
\And 
Giannis Daras \\ 
CS, UT Austin \\
\texttt{giannisdaras@utexas.edu} 
\AND
Eric Price \\ 
CS, UT Austin \\
\texttt{ecprice@cs.utexas.edu} 
\And
Alexandros G. Dimakis \\ 
ECE, UT Austin \\ 
\texttt{dimakis@austin.utexas.edu} 
\And 
Jonathan I. Tamir \\ 
ECE, UT Austin \\ 
\texttt{jtamir@utexas.edu} 
}

\begin{document}

\maketitle

\begin{abstract} 
The CSGM framework (Bora-Jalal-Price-Dimakis'17) has shown that deep
generative priors can be powerful tools for solving inverse problems.
However, to date this framework has been empirically successful only on
certain datasets (for example, human faces and MNIST digits), and it
is known to perform poorly on out-of-distribution samples. In this
paper, we present the first successful application of the CSGM
framework on clinical MRI data. We train a generative prior on brain
scans from the fastMRI dataset, and show that posterior sampling via
Langevin dynamics achieves high quality reconstructions. Furthermore,
our experiments and theory show that posterior sampling is robust to
changes in the ground-truth distribution and measurement process.
Our code and models are available at: 
\url{https://github.com/utcsilab/csgm-mri-langevin}.
\end{abstract}

\section{Introduction}

Compressed sensing~\cite{donoho2006compressed,candes2008restricted}
has enabled reductions to the number of measurements needed for
successful reconstruction in a variety of imaging inverse problems. In
particular, it has led to shorter scan times for magnetic resonance
imaging (MRI) \cite{lustig2007sparse,vasanawala2010csmri}, and most
MRI vendors have released products leveraging this framework to
accelerate clinical workflows.  Despite their successes,
sparsity-based methods are limited by the achievable acceleration
rates, as the sparsity assumptions are either hand-crafted or are
limited to simple learned sparse codes
\cite{bresler2011dictionarylearning,ravishankar2017datadriven}. 

More recently, deep learning techniques have been used as powerful
data-driven reconstruction methods for inverse problems
\cite{unser2017deepinverse,ongie2020deep}. There are two broad
families of deep learning inversion techniques~\cite{ongie2020deep}:
end-to-end supervised and distribution-learning approaches.
End-to-end supervised techniques use a training set of measured images
and deploy convolutional neural networks (CNNs) and other
architectures to learn the inverse mapping from measurements to image.
Network architectures that include both CNN blocks and the imaging
forward model have grown in popularity, as they combine deep learning
with the compressed sensing optimization framework, see
e.g.~\cite{hammernik2018learning,aggarwal2018modl,mardani2018deep}.
End-to-end methods are trained for specific imaging anatomy and
measurement models and show excellent performance in these tasks.
However, reconstruction quality is known to suffer when applied out of
distribution, and recently has been shown to severely
degrade~\cite{antun2020instabilities,darestani2021measuring} under
certain types of natural measurement and anatomy perturbations. 

In this paper we study deep learning inversion techniques based on
distribution learning.  These models are trained without reference to
measurements, and so easily adapt to changes in the measurement
process.  The most common family of such techniques, known also as
Compressed Sensing with Generative Models
(CSGM)~\cite{bora2017compressed} uses pre-trained generative models as
priors.  Generative models are extremely powerful at representing
image statistics and CSGM has been successfully applied to numerous
inverse problems~\cite{bora2017compressed,hand2019global} including
non-linear phase retrieval~\cite{hand2018phase}, and improved with
invertible models~\cite{asim2019invertible}, sparsity based
deviations~\cite{dhar2018modeling}, image
adaptivity~\cite{hussein2020image}, and posterior
sampling~\cite{song2019generative, jalal2021instance}.  These methods
have only recently been applied to MRI and have not yet been shown to
be competitive with supervised end-to-end methods.  The very recent
work \cite{kelkar2021prior} trains a StyleGAN for magnitude-only DICOM
images but requires the presence of side-information and studies
Gaussian, real-valued measurements for reconstruction.  The deviation
from the true MRI measurement model and the use of magnitude images
are known to be problematic when evaluating performance
\cite{shimron2021subtle}.  Another work~\cite{kelkar2021compressible}
trained an Invertible Neural Network on complex-valued single-coil MR
images and showed very good performance in comparison to sparsity and
GAN priors.  Untrained and unamortized
generators~\cite{heckel2018deep} have also been recently explored
\cite{darestani2021measuring}, showing promising results in some
cases. Further, \cite{cole2021fast} studies the harder problem of
learning a generative model for a class of images using only partial
observations, as first proposed in
AmbientGAN~\cite{bora2018ambientgan}.

In this paper we train the first score-based generative
model~\cite{song2020improved} for MR images. We show that we can
faithfully represent MR images without any assumptions on the
measurement system. As a consequence, we are able to reconstruct
retrospectively under-sampled MRI data under a variety of realistic
sampling schemes. We show that our reconstruction algorithm is
competitive with end-to-end supervised training when the test-data are
matched to the training data and that it is robust to various
out-of-distribution shifts, while in some cases end-to-end methods
significantly degrade.

\begin{figure}[t]
    \centering
    \includegraphics[width=\linewidth]{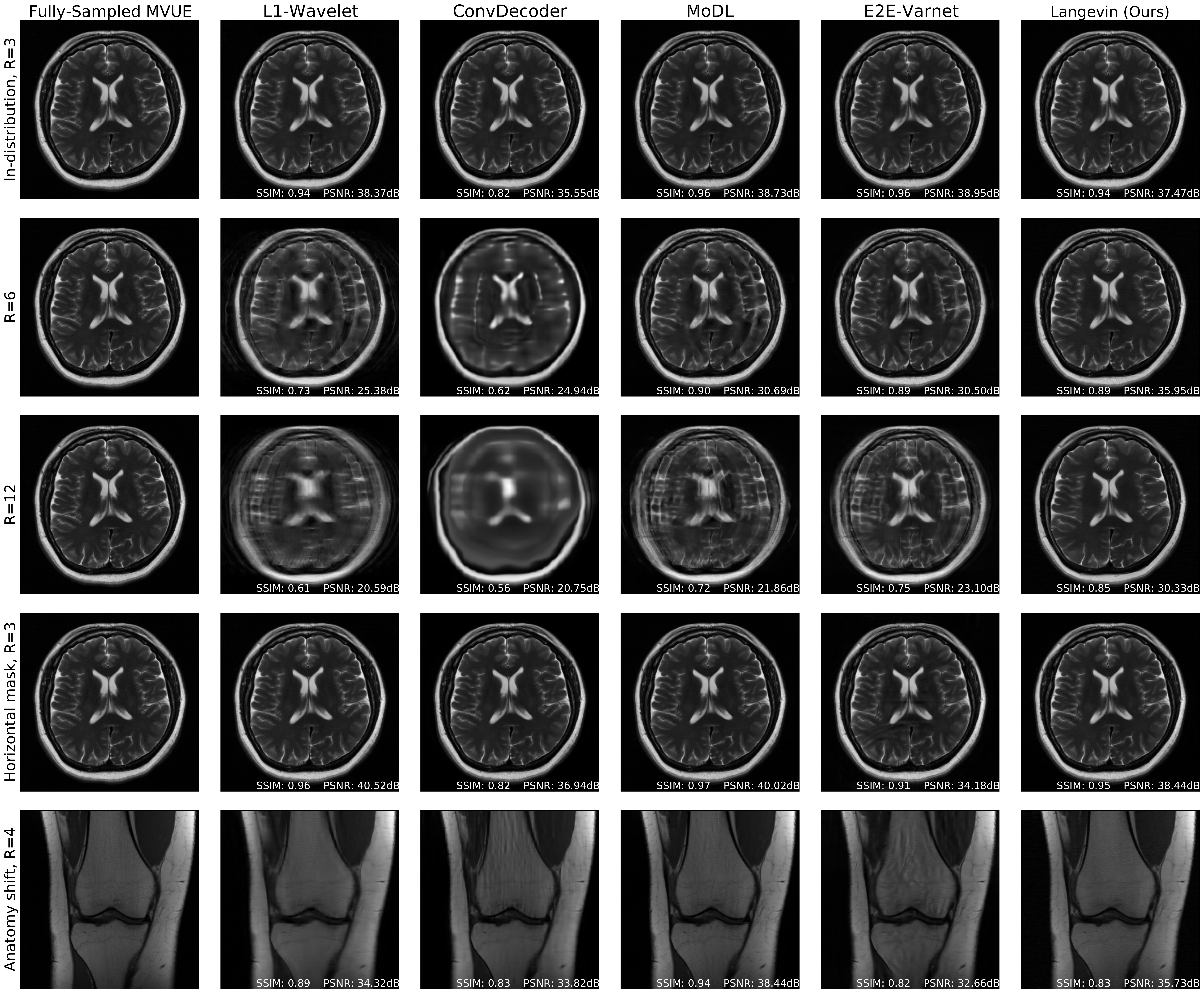}
    \caption{\small Comparison of reconstruction methods for
        in-distribution, sampling-shift, and anatomy-shift images.
        All methods and hyperparameters were optimized on T2-weighted
        \emph{brain} scans with a vertical sampling mask, and tested
        at higher accelerations, horizontal masks, and on knee \&
        abdomen scans.  Our reconstructions are competitive with
        state-of-the-art methods, and introduce fewer artifacts out of
        distribution.  All measurements are multicoil k-space from the
        NYU fastMRI dataset and the supervised baselines are trained
        from scratch on MVUE targets for a fair comparison.}
    \label{fig:main}
\end{figure}

\subsection{Contributions}
\begin{itemize}[leftmargin = *]

    \item We successfully train a score-based deep generative model
        for complex-valued, T2-weighted brain MR images without any
        assumptions on the measurement scheme. When applied to
        multi-coil MRI reconstruction under the CSGM framework, we
        achieve competitive performance compared to end-to-end deep
        learning methods when the test-time data are sampled within
        distribution.
    \item We give evidence that posterior sampling should give
        high-quality reconstructions.  First, we show that for any
        measurements (including the Fourier measurements in MRI) that
        posterior sampling with the correct prior is within constant
        factors of the optimal recovery method; second, even if the
        prior is wrong but gives $\alpha$ mass to the true
        distribution, we show that posterior sampling for Gaussian
        measurements is nearly optimal with just an additive
        $O(\log(1/\alpha))$ loss.
    \item We empirically show that our approach is robust to test-time
        distribution shifts including different sampling patterns and
        imaging anatomy.  The former is unsurprising given that our
        model was trained without knowledge of the measurement scheme.
        As a consequence, our approach provides a degree of
        flexibility in choosing scan parameters -- a common situation
        in routine clinical imaging. Perhaps surprisingly, the latter
        indicates that a specialized training set may offer sufficient
        regularization for a larger class of images.  In contrast, we
        empirically show that end-to-end methods do not always enjoy
        the same robustness guarantees, in some cases leading to
        severe degradation in reconstruction quality when applied
        out-of-distribution. 
    \item Our method can be used to obtain multiple samples from the
        posterior by running Langevin dynamics with different random
        initializations. This allows us to get multiple
        reconstructions which can be used to obtain confidence
        intervals for each reconstructed voxel and visualize our
        reconstruction uncertainty on a voxel-by-voxel resolution.
        Uncertainty quantification can be incorporated into end-to-end
        methods, e.g., using variational auto-encoders
        \cite{mardani2021mriuncertainty}, but this requires changes to
        the architecture. Our method does not require any modification
        and multiple reconstruction samplers can be run in parallel.
\end{itemize}
        
Our main results are succinctly summarized in Figure~\ref{fig:main}:
we achieve equivalent reconstruction performance using a reduced
training set when evaluated in-distribution and are robust when
evaluated out-of-distribution.

\subsection{Related Work} 

Generative priors have shown great utility to improving compressed
sensing and other inverse problems, starting
with~\cite{bora2017compressed}, who generalized the theoretical
framework of compressed sensing and restricted eigenvalue
conditions~\cite{tibshirani1996regression,
donoho2006compressed,bickel2009simultaneous, candes2008restricted,
hegde2008random, baraniuk2009random, baraniuk2010model,
eldar2009robust} for signals lying on the range of a deep generative
model~\cite{goodfellow2014generative, kingma2013auto,
song2021scorebased}.  Lower bounds in~\cite{kamath2019lower,
liu2019information, jalali2019solving} established that the sample
complexities in~\cite{bora2017compressed} are order optimal.  The
approach in~\cite{bora2017compressed} has been generalized to tackle
different inverse problems~\cite{jalal2020robust, hand2018phase,
    asim2018solving, qiu2019robust, liu2020sample, mardani2017deep,
    rick2017one, balevi2020high}, and different reconstruction
    algorithms~\cite{dhar2018modeling, kabkab2018task,
    pandit2019inference, fletcher2018inference, fletcher2018plug,
mardani2018deep, heckel2018deep, heckel2020compressive,
daras2021intermediate}.  The complexity of optimization algorithms
using generative models have been analyzed in~\cite{gomez2019fast,
hegde2018algorithmic, lei2019inverting, hand2017global}.  Our prior
work shows that posterior sampling is instance-optimal for compressed
sensing~\cite{jalal2021instance}, and satisfies certain fairness
guarantees without explicit information about protected sensitive
groups~\cite{jalal2021fairness}.

Using compressed sensing for multi-coil MRI reconstruction has led to
a rich body of work in the past two decades \cite{lustig2007sparse,
deshmane2012parallel, uecker2014espirit, rosenzweig2018simultaneous}.
See~\cite{doneva2021csmri} and the recent special
issue~\cite{doneva2020ieeespsguest} for an overview of these methods.
Classical approaches impose sparsity in a well-chosen basis, such as
the wavelet domain \cite{lustig2007sparse}, or apply shallow learning
that leverages low-level redundancy in the images
\cite{bresler2011dictionarylearning, ravishankar2017datadriven,
bresler2020ieeesps}.  Recent research has demonstrated the superior
performance of deep neural networks for MR image reconstruction
\cite{schlemper2017deep, hammernik2018learning,
aggarwal2018modl,sriram2020end, Sriram_2020_CVPR}.  A broad class of
approaches is represented by end-to-end unrolled methods, which use
deep networks as learned data priors in the image
\cite{aggarwal2018modl, hammernik2018learning, sriram2020end} or
k-space domain \cite{sriram2020grappanet}. Recent work has also
investigated the performance of untrained methods
\cite{ulyanov2018deep, heckel2020compressive} for MR reconstruction
and has shown competitive results. A much less explored line of
research is MR image reconstruction with generative priors. The work
in \cite{narnhofer2019inverse} proposes a CSGM-like algorithm that
finetunes an entire pre-trained generator that requires a carefully
tuned optimization algorithm during inference.

\begin{figure}[t]
    \centering
    \includegraphics[width=\columnwidth]{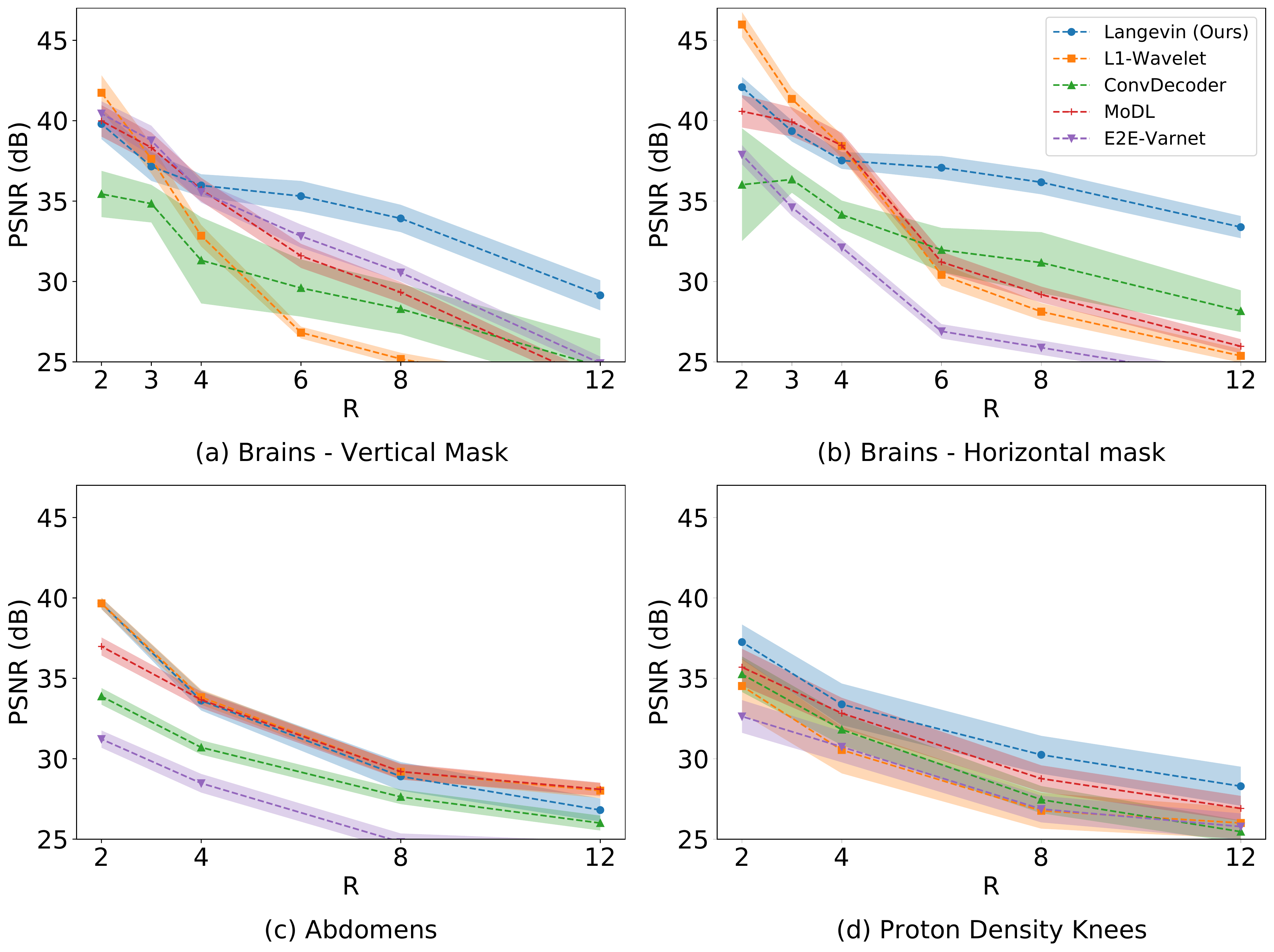}
    \caption{\small Average test PSNR in various scenarios, across a range of acceleration factors $R$. Higher $R$ indicates a smaller number of acquired measurements. All methods and hyperparameters were optimized on brains with an equispaced vertical mask. Our approach mostly shows the best performance and lowest reconstruction variance both in- and out-of-distribution at test-time. Shaded regions indicate $95\%$ confidence intervals. Note that we trained baselines on MVUE images and hence these numerical values should not be compared with those in literature trained on RSS images (see Appendix~\ref{app:mvue-rss} for a more detailed discussion).}
    \label{fig:main-psnr}
\end{figure}

\section{System Model and Algorithm}

\subsection{Multi-coil Magnetic Resonance Imaging}
MRI is a medical imaging modality that makes measurements using an array of radio-frequency coils placed around the body. Each coil is spatially sensitive to a local region, and measurements are acquired directly in the spatial frequency, or \textit{k-space}, domain.
To decrease scan time, reduce operating costs, and improve patient comfort, a reduced number of k-space measurements are acquired in clinical use and reconstructed by incorporating explicit or implicit knowledge of the spatial sensitivity maps \cite{sodickson1997simultaneous,pruessmann1999sense,griswold2002grappa}. 
Formally, the vector of measurements $y_i \in \C^L$ acquired by the \nth{i} coil can be characterized by the forward model \cite{pruessmann1999sense}:
\begin{equation}
    y_i = P F S_i x^* + w_i,\quad i=1,...,N_c,
    \label{eq:mri_model}
\end{equation}
\noindent where $x^* \in \C^N$  is the image containing $N$ pixels, $S_i$ is an operator representing the point-wise multiplication of the \nth{i} coil sensitivity map, $F$ is the spatial Fourier transform operator, $P$ represents the k-space sampling operator, and we assume $w_i \sim \mathcal{N}_c\left(0, \sigma^2I\right)$ for simplicity. Importantly, note that the same under-sampling operator is applied to all $N_c$ coils. 

The acceleration factor $R$ denotes the degree of under-sampling in the $k$-space domain, i.e., $R = N / L$.  
Due to the multiple coils, the measurements may not be compressive for small $R$. However, due to redundancy between the coils, the measurements are compressive for moderate values of $R$ (even if $N_c \cdot L > N$)~\cite{huang2008software}.
Also note that we use the \emph{true acceleration factor} $R$, and this does not match the values in fastMRI~\cite{knoll2020fastmri}~\footnote{\url{https://github.com/facebookresearch/fastMRI/blob/main/fastmri/data/subsample.py}, line 247 has the fastMRI definition of equispaced acceleration factors.} on certain sampling patterns.

Given multi-coil measurements $y$, sensitivity maps represented by $S$ and the sampling operator $P$, the goal of MR image reconstruction is to estimate the underlying image variable $x^*$. Prior work formulates this as a regularized optimization problem:
\begin{equation}
    \argmin_{x} \norm{y - A x}_2^2 + \lambda Q(x),
    \label{eq:mri_recon}
\end{equation}
\noindent where we use the operator $A\in\C^{M\times N} (\text{ with } M=N_c \cdot L)$ to subsume the discrete approximation to all linear effects, and $Q$ is a suitably chosen functional prior for the image variable $x$. For example, to enforce a sparsity prior, one can penalize the $\ell_1$ norm in the wavelet representation of $x$ \cite{lustig2007sparse}. More recent approaches involve learned regularization terms parameterized by deep neural networks \cite{schlemper2017deep,hammernik2018learning,aggarwal2018modl}. These models are typically trained \textit{end-to-end} using a fixed training set and certain assumptions about the sampling operator. In the sequel, we present how score-based generative models can be combined with the posterior sampling~\cite{jalal2021instance} mechanism to reformulate \eqref{eq:mri_recon} and achieve good quality reconstructions without any \textit{a priori} assumptions about the sampling scheme. 

When k-space is fully sampled at the Nyquist rate and no regularization is applied, the solution to \eqref{eq:mri_recon} corresponds to the minimum-variance unbiased estimator (MVUE) of $x^*$, denoted by $\hat{x}_{\mathrm{MVUE}}$ \cite{pruessmann1999sense}. Given fully sampled k-space data, this estimate can act as a reference image for evaluating reconstruction error as well as for end-to-end training.
Alternatively, a reference image called the root-sum-of-squares (RSS) estimate can be formed
by taking the inverse Fourier transform of each coil and subsequently applying the $\ell_2$ norm for each pixel across the coil dimension, i.e. $\hat{x}_{\mathrm{RSS}} =\sqrt{\sum_{i=1}^{N_c} \left|\left(F^H y_i\right)\right|^2}$,
where $F^H$ is the Hermitian transpose of $F$ (here the inverse DFT).
Although the RSS estimate is a biased estimator, it is often used as it does not make any assumptions about the sensitivity maps, which are not explicitly measured by the MRI system.
However, even if solving \eqref{eq:mri_recon} results in perfect recovery of $x^*$, there will be a bias when comparing the result to $\hat{x}_{\mathrm{RSS}}$ and thus the RSS and MVUE cannot be directly compared numerically.

\subsection{Posterior Sampling}

The algorithm we consider is \emph{posterior
sampling}~\cite{jalal2021instance}.  That is, given an observation of
the form $y = Ax^* + w$, where $y \in \C^M$, $A \in \C^{M \times N}$,
$w \sim \cN_c(0,\sigma^2 I)$, and $x^* \sim \mu$, the posterior
sampling recovery algorithm outputs $\xhat$ according to the posterior
distribution $\mu(\cdot | y)$. 

In order to sample from the posterior, we use \emph{Langevin
Dynamics}~\cite{bakry1985diffusions}. Assuming we have access to
$\nabla_x \log \mu(x|y)$, we can sample from $\mu(x|y)$ by running
noisy gradient ascent:
\begin{align}
  x_{t+1} \leftarrow x_t + \eta_t \nabla_{x_t} \log \mu(x_t |
  y) + \sqrt{2 \eta_t} \; \zeta_t, \quad \zeta_t \sim \cN(0,1).
  \label{eqn:langevin}
\end{align}
Prior work~\cite{bakry1985diffusions} has shown that as $t\to \infty$
and $\eta_t \to 0$, Langevin dynamics will correctly sample from
$\mu(x|y)$.  In practice, vanilla Langevin Dynamics are slow to
converge. Hence, the work in~\cite{song2019generative} proposes
\emph{annealed} Langevin Dynamics, where the marginal distribution of
$x$ at iteration $t$ is modelled as $\mu_t = \mu * \cN(0, \beta_t^2)$
and the generative model is trained to estimate the score function
$f(x_t; \beta_t) := \nabla_{x_t} \log ((\mu * \cN(0,\beta_t^2)(x_t))$.

Since the distribution of $y|x^*$ is Gaussian in
Eqn~\eqref{eq:mri_recon}, we obtain $\nabla_{x_t}\log \mu(y | x_t) =
\frac{A^H (y- A x_t) }{\sigma^2}$.  We find that it is also helpful to
anneal this term, and we set it to $\frac{A^H (y - Ax_t)}{\sigma^2 +
\gamma_t^2}$, where $\gamma_t\to 0$ is a decreasing sequence. An
application of Bayes' rule gives: $\nabla_{x_t} \log \mu(x_t | y) =
f(x_t; \beta_t) + \frac{A^H (y - Ax_t)}{\sigma^2 + \gamma_t^2}$.

Putting everything together, our final algorithm is: for $x_0 \sim
\cN_c(0,I)$ and for all $t=0, \cdots, T-1$, 
\begin{align}
  x_{t+1} \leftarrow x_t + \eta_t \left(f(x_t; \beta_t) + \frac{A^H (y - A
  x_t)}{\gamma^2_t + \sigma^2}\right) + \sqrt{2 \eta_t} \; \zeta_t,
	\quad \zeta_t
  \sim \cN(0; I).
  \label{eqn:annealed-langevin}
\end{align}
Note that the parameters $T, \{ \beta_t\}_{t=0}^{T-1}$ were fixed
during training of the generative model, and hence the only
hyperparameters during inference are $\{\eta_t\}_{t=0}^{T-1}, \sigma$ and
$\{\gamma_t\}_{t=0}^{T-1}$. Scripts in our codebase
describe hyperparameter values used in our experiments.

\section{Theoretical Results}\label{sec:theory}
\paragraph{Background and Notation.}
We first introduce background and notation required for our theoretical results. $\norm{\cdot}$ refers to the $\ell_2$ norm. In this section alone, for simplicity of exposition, we will assume that all matrices and vectors are real valued.  

For two probability distributions $\mu, \nu$ on some normed space $\Omega$, and for any $q\geq 1$, the Wasserstein-$q$~\cite{villani2008optimal, arjovsky2017wasserstein} and Wasserstein-$\infty$~\cite{champion2008wasserstein} distances are defined as:
\begin{align*}
  \cW_q ( \mu, \nu) := \inf_{\gamma \in \Pi(\mu,\nu)} \left( \E_{(u,v)
  \sim \gamma} \left[ \norm{ u - v}^q \right] \right)^{1/q}, \quad
  \cW_{\infty}(\mu,\nu) := \inf_{\gamma \in \Pi(\mu,\nu)} \left(
    \underset{{(u,v) \in \Omega^2}}{\gamma\text{-}\esssup} \norm{u - v}
\right).
\end{align*}
where $\Pi(\mu,\nu)$ denotes the set of joint distributions whose
marginals are $\mu,\nu$.
The above definition says that if $\cW_{\infty}(\mu,\nu)\leq \eps$,
and $(u,v) \sim \gamma$, then $\norm{u-v} \leq \eps$ almost surely.

The $(\eps, \delta)-$\emph{approximate
covering number}~\cite{jalal2021instance}, is defined as the smallest
number of $\eps$-radius balls required to cover $1-\delta$ mass under
a distribution.  
\begin{definition}[$(\eps,\delta)$-approximate covering number]
  Let $\mu$ be a distribution on $\R^N$. For some parameters
  $\eps > 0,
  \delta\in \left[ 0,1 \right],$ the \emph{$(\eps,\delta)$-approximate
  covering number} of $\mu$ is defined as
  \[
    \cov_{\eps,\delta}(\mu) := \min\left\{ k :  \mu\left[ \cup_{i=1}^k
    B(x_i,\eps)  \right]\geq 1 - \delta, x_i \in \R^N  \right\},
  \]
  where $B(x, \eps)$ is the $\ell_2$ ball of radius $\eps$ centered
  at $x$.
\end{definition}\label{defn: cov}

\paragraph{Distributional robustness under Gaussian measurements.}
First, we consider mismatch between the ground-truth distribution, denoted by $\mu$, and the generator distribution, denoted by $\nu$. Prior work~\cite{jalal2021instance} has shown that if (i) $\cW_q(\mu, \nu) \leq \eps$ for some $q\geq 1$ and (ii) we are given $M \geq O(\log \cov_{\eps, \delta} (\mu))$ Gaussian measurements, then posterior sampling with respect to $\nu$ will recover $x^* \sim \mu$ up to an error of $\eps / \delta^{1/q}$ with probability $1-\delta$. Closeness in Wasserstein distance is a reasonable assumption in certain examples, such as when $\mu$ is the distribution of celebrity faces and $\nu$ is the distribution of a generator trained on FlickrFaces \cite{karras2019style}. However, this assumption is unsatisfactory when we consider distributions of abdominal and brain MR scans, for example, since images of these anatomies look entirely different.  

We define the following weaker notion of divergence between distributions. Informally, this new definition tells us that $\nu$ and $\mu$ are ``close'' if they can each be split into components which are close in $\cW_\infty$ distance, such that the close components contain a sufficiently large fraction under $\nu$ and $\mu$. Formally, this is defined as:
\begin{definition}[$(\delta,\alpha)\text{-}\cW_\infty$ divergence]
  For two probability distributions $\nu$ and $\mu$, and parameters
  $\delta, \alpha \in [0,1]$, the $(\delta, \alpha)\text{-}\cW_\infty$
  divergence is defined as
  \begin{align*}
    &(\delta, \alpha)\text{-}\cW_\infty(\mu, \nu) := \inf \{ \eps \geq 0: \\ 
    & \exists \mu', \mu'', \nu', \nu'' \in \cM(\R^N) \; s.t.\; \mu = (1-\delta)
  \mu' + \delta \mu'', \nu = (1 - \alpha) \nu' + \alpha \nu'', \cW_\infty(\mu',
\nu') = \eps. \}
  \end{align*}
\end{definition}
Lemma~\ref{lemma:winf} highlights that this is a strict generalization of Wasserstein distances, in the sense that closeness in Wasserstein distance implies closeness in this new divergence.

Since the $(\delta,\alpha)\text{-}\cW_\infty$ divergence is a
generalization of Wasserstein distances, it is not clear that the main
Theorem in~\cite{jalal2021instance} holds for distributions that are close in
this new divergence.  The following result shows a rather
surprising fact: if $(\delta,\alpha)\text{-}\cW_\infty(\mu , \nu) \leq \eps$
then posterior sampling with $M =O \left( \log \left( \frac{1}{1 -
\alpha} \right) + \log \cov_{\eps,\delta} (\mu) \right) $
measurements will still succeed with probability $\geq 1-O(\delta)$.  

\define{thm: main}{Theorem}{%
Let $\delta,\alpha \in [0,1]$, and $\eps>0$ be
parameters.  Let $\mu, \nu$ be arbitrary distributions over $\R^N$
satisfying $(\delta,\alpha)\text{-}\cW_\infty(\mu, \nu) \leq \eps$.  
Let $x^* \sim \mu$ and suppose $y=Ax^* + w$, where $A\in\R^{M\times
N}$ and $w\in\R^{M}$ are i.i.d. Gaussian normalized such that
$A_{ij} \sim \cN(0,1/M)$ and $w_i\sim \cN(0,\sigma^2/M)$, with
$\sigma \gtrsim \eps$.  Given $y$ and the fixed matrix $A$, let
$\xhat$ be the output of posterior sampling with respect to $\nu$. 

Then for $M\geq O\left( \log\left(\frac{1}{1-\alpha}\right) +
\min(\log \cov_{\sigma, \delta}( \mu ), \log \cov_{\sigma,
\delta} ( \nu ))\right)$, there exists a universal constant $c > 0$ such
that with probability at least $ 1 - e^{-\Omega(M)}$ over $A, w$,
\begin{align*}
\Pr_{x^*\sim \mu , \xhat \sim \nu( \cdot | y)} \left[ \norm{x^* - \xhat}
\geq c (\eps +  \sigma )\right] &  \leq \delta + e^{-\Omega(M)}.
\end{align*}
}%
\state{thm: main}%

For our running example of $\nu$ being a generator trained on brain scans, and $\mu$ the distribution of abdominal scans, we can set $\nu'$ to be the distribution of our generator restricted to abdominal scans, and we can let $\mu'$ be the distribution restricted to ``inliers'' in $\mu$. This shows that even if our generator places an \emph{exponentially small} probability mass(i.e., $1-\alpha \ll 1$) on the set of abdominal scans, we can still recover abdominal scans with a \emph{polynomial additive} increase in the number of measurements (i.e., $\log(1/(1-\alpha))$).

\paragraph{Near-optimality under arbitrary measurement processes.} The previous result required Gaussian matrices to handle the distribution shift. Our next result shows that for an \emph{arbitrary} measurement process, and assuming that there is no distribution shift between the generator and the ground truth distribution, posterior sampling is almost the best algorithm for this \emph{fixed}  measurement process.
This result also shows that posterior sampling is good with respect to \emph{any} metric.

\define{thm:optimal}{Theorem}
{
  Let $d(\cdot, \cdot)$ be an arbitrary metric over $\R^N \times \R^N$. Let $x^* \sim \mu$ and let $y=\cA(x^*)$ be measurements generated from
  $x^*$ for some arbitrary forward operator $\cA:\R^N \to \R^M$. Then
  if there exists an algorithm that uses $y$ as inputs and outputs
  $x'$ such that 
  $$d(x^*, x') \leq \eps \text{ with probability } 1 - \delta,$$
  then posterior sampling $\xhat \sim \mu(\cdot | y)$ will satisfy
  $$d(x^* , \xhat) \leq 2\eps \text{ with probability } \geq 1 -
  2\delta.$$
}
\state{thm:optimal}

\paragraph{Remark on combining these results.}  Our theoretical results above show that posterior sampling is (1) highly robust to distribution shift under Gaussian measurements, and (2) accurate with arbitrary measurements without distribution shift.  A natural hope would be to combine these two results and show that it is robust to distribution shift under Fourier measurements.  Unfortunately, this is \emph{not} true for general distributions: for example, if $\mu$ and $\nu$ are both random distributions over Fourier-sparse signals, then Fourier measurements will usually give zero information about the signal, so cannot convince the sampler to sample near $\mu$ rather than $\nu$.

\section{Experimental Results}\label{sec:exp}
We perform retrospective under-sampling in all experiments, i.e., given fully-sampled k-space measurements from the NYU fastMRI~\cite{knoll2020fastmri,zbontar2018fastMRI} and Stanford MRI~\cite{mridataorg} datasets, we apply sampling masks and evaluate the performance of all considered algorithms on the reconstructed data. Depending on scan parameters (e.g., 3D scans for the Stanford knee data in Appendix~\ref{app:3d-knees}), we appropriately slice and sample the data in the proper dimension so as to not commit any inverse crime~\cite{guerquin2011realistic,shimron2021subtle}.

\begin{figure}
\begin{center}
    \includegraphics[width=\columnwidth]{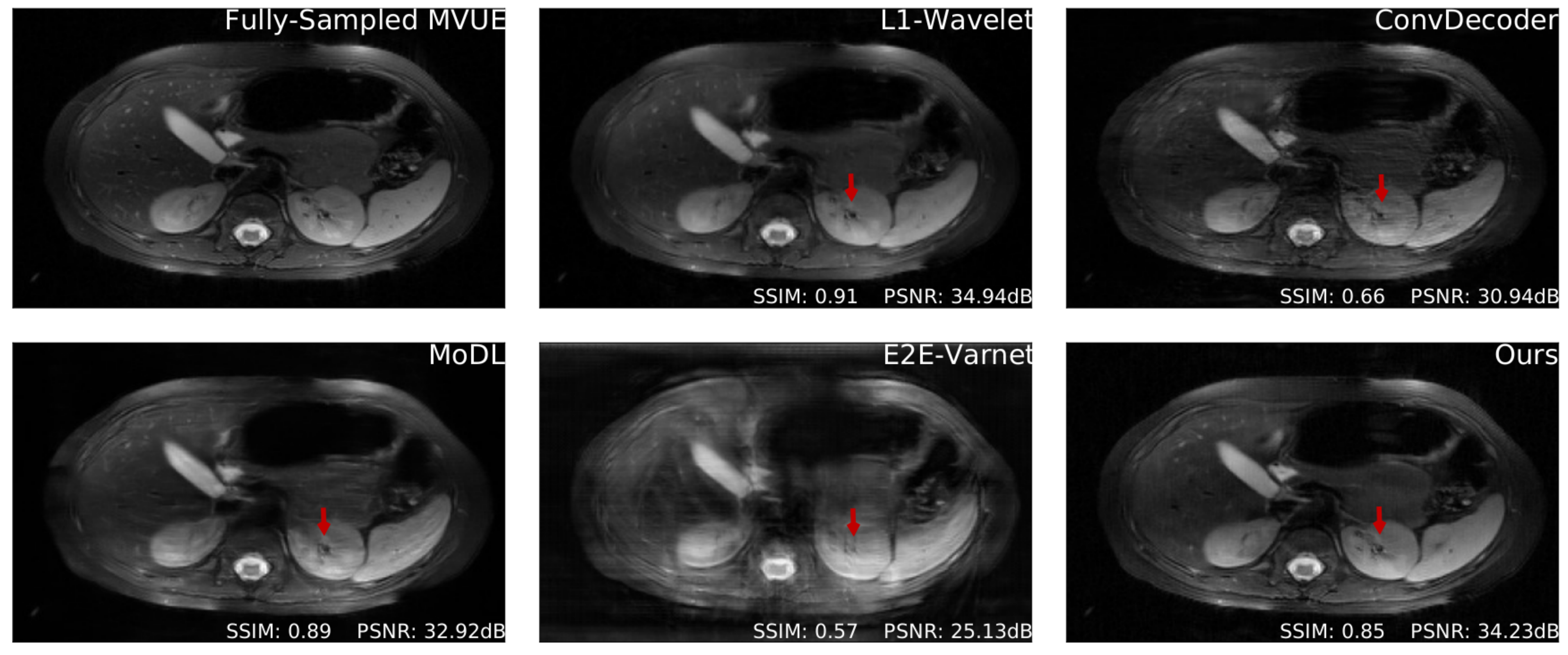}
\end{center}
\caption{\small Comparative reconstructions of a 2D abdominal scan
    with uniform random under-sampling in the horizontal direction at
    $R=4$. None of the methods were trained to reconstruct abdomen
    MRI. Our method uses a score-based generative model trained on
    brain images (as explained) and obtains good reconstructions. The
    red arrows indicate missing details or artifacts in the kidney
    structure. 
}
\label{fig:abdomen-equi-horizontal-recons}
\end{figure}

We first highlight that an advantage of the proposed approach is the
invariance to the sampling scheme during training. In contrast, this
is a design choice that must be made for supervised end-to-end
methods, which here were trained on equispaced, vertical sampling
masks, following the fastMRI 2020 challenge guidelines
\cite{zbontar2018fastMRI,muckleyfastmri2020}. As our results show,
this affords us a significant degree of robustness across a wide
distribution of sampling masks during inference.

We train a score-based model, NCSNv2~\cite{song2020improved}, on a
small subset of scans from the NYU fastMRI brain dataset.
Specifically, we train using T2-weighted images at a field strength of
3 Tesla for a total of 14,539 2D training slices.  We calculate the
MVUE from the fully sampled data and use the ESPIRiT algorithm
\cite{uecker2014espirit,iyer2020autoespirit} applied to the
fully-sampled central portion of k-space to estimate the sensitivity
maps.  The backbone network for our model is a
RefineNet~\cite{lin2017refinenet}.  Since the generator's output is
expected to be complex-valued, we treat the real and imaginary parts
as separate image channels. Details about the architectures are given
in Appendix~\ref{app:implementation}.

We use an $\ell_1$-Wavelet regularized reconstruction algorithm
\cite{lustig2007sparse} as a parallel imaging and compressed sensing
baseline.  This aims to solve the optimization problem given in
\eqref{eq:mri_recon} with $Q(x) = ||Wx||_1$, where $W$ is a 2D Wavelet
transform.  We use the publicly available implementation from the BART
toolbox \cite{uecker2015berkeley,bart} and optimize the regularization
hyper-parameter using the same subset of samples from the brain
dataset that was used to train our method. We find that $\lambda =
0.01$ performs the best on the training data and use this value for
all experiments.  We consider three different deep learning baselines:
MoDL \cite{aggarwal2018modl}, E2E-VarNet \cite{sriram2020end}, and the
ConvDecoder architecture \cite{darestani2021measuring}. 

We train the MoDL and E2E-VarNet baselines \textit{from scratch} on
the same training dataset as our method, at acceleration factors
$R=\{3,6\}$ and equispaced under-sampling, with a supervised SSIM loss
on the magnitude MVUE image, for $40$ and $15$ epochs, respectively,
using a batch size of $1$. For the ConvDecoder baseline, we use the
architecture for brain data in \cite{darestani2021measuring} that
outputs a complex image estimate and optimize the number of fitting
iterations on a subset of samples from the training data. We find that
$10000$ iterations are sufficient to reach a stable average
performance at $R=3$. Put together, all of our baselines are tailored
to estimate the complex image $x$, thus all comparisons are fair. We
evaluate reconstruction performance using the complex MVUE of the
fully sampled data as a reference image and measure the peak
signal-to-noise ratio (PSNR) and structural similarity index (SSIM)
\cite{wang2006modern} between the absolute values of the
reconstruction and ground-truth MVUE images.

\subsection{In-Distribution Performance}
In this experiment, we test all models using the same forward model that matches the training conditions for the baselines: vertical, equispaced sampling patterns. Examples of various sampling patterns are shown in Appendix~\ref{app:brains}.

Figure~\ref{fig:main} (top three rows) shows qualitative results and Figures~\ref{fig:main-psnr}a \& \ref{fig:main-ssim}a respectively show PSNR \& SSIM values, for the case where there is no mismatch between the training and inference sampling patterns.
As the baselines were trained to maximize SSIM at $R=3\;\&\;6$, we see that they achieve better SSIM scores than us at these accelerations, although there is clear aliasing in the baselines at $R=6$. We achieve better PSNR values at these accelerations, which supports the claim that our method does not overfit to a particular metric (Theorem~\ref{thm:optimal}).
This also highlights the importance of qualitative evaluations in medical image reconstruction and the limitations of existing image quality metrics \cite{mason2019comparison}.
From the third row of Figure~\ref{fig:main}, and Figures~\ref{fig:main-psnr}a \& \ref{fig:main-ssim}a, we notice that our method surpasses baselines at higher accelerations. 

We find that $\ell_1$-Wavelet suffers both qualitatively and quantitatively at high acceleration factors, while the ConvDecoder is also a competitive architecture, but incurs a large computational cost. When benchmarked on an NVIDIA RTX 2080Ti GPU, our method takes $16$ minutes and $0.95$ GB of memory to reconstruct a high-resolution brain scan, whereas the ConvDecoder takes longer than $80$ minutes and $6.6$ GB of memory. While our method is limited by the inference time and is not in the range of end-to-end models (where reconstruction takes at most on the order of seconds and $3.5$ GB of memory), multiple scans can be reconstructed in parallel due to the reduced memory footprint.
\begin{figure}[t]
\begin{center}
  \includegraphics[width=\columnwidth]{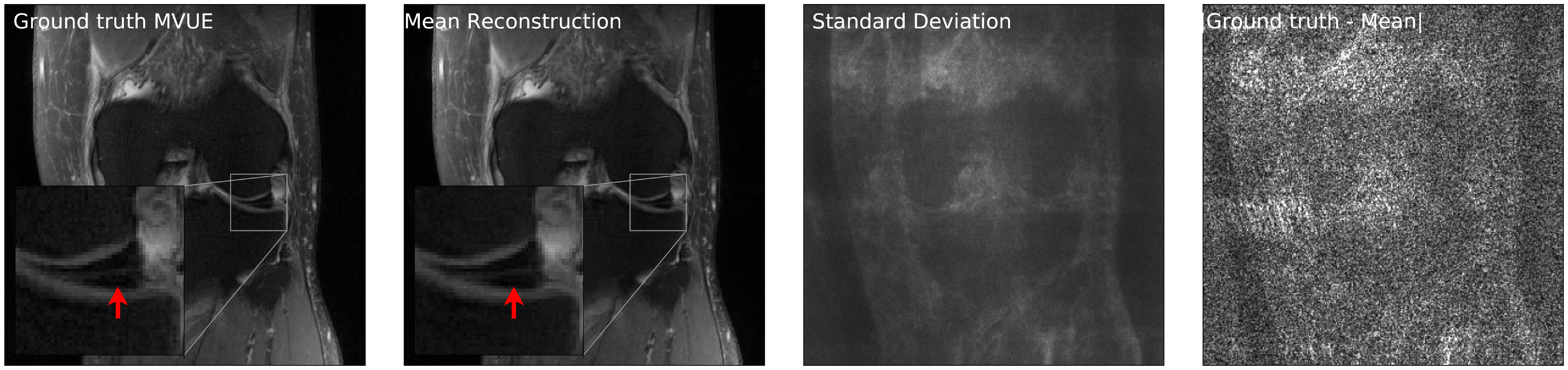}
\end{center}
\caption{\small Our method successfully recovers fine details and can provide an estimate of the reconstruction error. The left column shows a knee from the fastMRI dataset, along with an annotated meniscus tear (indicated by red arrow in zoomed inset). Given measurements at an acceleration factor of $R=4$, we obtain $48$ independent reconstructions via posterior sampling. The second column shows the pixel-wise average of reconstructions, the third column shows the pixel-wise standard deviation, and the fourth column shows the magnitude of the error between the ground truth and the mean reconstruction. Note that our generative prior has never seen such pathology, as it was trained on T2-weighted brain scans.}
\label{fig:tear}
\end{figure}

\subsection{Out-of-Distribution Performance}\label{sec:exp-out}

\paragraph{Test-time sampling pattern shifts.}
Here we consider shifts in the forward sampling operator at test-time, while still evaluating on the same anatomy as the training conditions. We measure robustness by evaluating the average incurred performance loss when the sampling pattern changes. Recall that our proposed approach does not use any explicit information about the sampling pattern $P$ during training, hence we anticipate the highest degree of robustness. 

Figure~\ref{fig:main} (fourth row) shows qualitative reconstructions when the measurements are obtained from an equispaced, horizontal sampling mask, with an acceleration factor $R=3$. It can be observed that the reconstructions output by E2E-VarNet show aliasing artifacts. Based on the statistical results in Figure~\ref{fig:main-psnr}b \& \ref{fig:main-ssim}b, our method retains its performance.

Furthermore, this experiment reveals that MoDL is more robust to this type of mask shift when compared to E2E-VarNet, even though it uses a smaller network. This is explained by the fact that E2E-VarNet does not use external sensitivity map estimates, but uses a deep neural network for end-to-end map estimation. While this improves performance on in-distribution samples, the performance drop is strong evidence that accurate sensitivity map estimation is vital for robust generalization, and both our proposed approach and MoDL benefit from the external ESPIRiT algorithm, which is compatible with different sampling patterns.

We do note that retrospectively flipping the horizontal and vertical sampling direction is not necessarily representative of prospective sampling in the horizontal direction due to the discrete nature of the phase encoding direction in MRI, and this may contribute to the higher scores compared to the vertical mask experiments.

\paragraph{Test-time anatomy shifts.}
We now consider the more difficult problem of reconstructing different anatomies than the ones seen during. This was previously investigated in \cite{darestani2021measuring}, which concluded that all methods suffer a drastic shift due to the various changes in scan parameters between body parts. In contrast to prior work, our main finding is that the proposed score-based model retains a significant degree of robustness under these shifts, and outputs excellent qualitative reconstructions. In some cases, some end-to-end methods retain robustness as well.

Figures~\ref{fig:main-psnr}c \& \ref{fig:main-ssim}c show PSNR and SSIM scores obtained on reconstructed abdominal scans obtained from \cite{mridataorg} at different acceleration factors. This represents both an anatomy and sampling pattern shift, and it can be seen that our method, MoDL, and the $\ell_1$-Wavelet algorithm retain their competitive advantage, while the ConvDecoder and E2E-VarNet suffer severe performance losses. Figure~\ref{fig:abdomen-equi-horizontal-recons} further shows a qualitative comparison of a reconstructed abdominal scan at $R=4$, with highlighted artifacts. Appendix~\ref{app:abdomens} shows another abdomen scan.

Finally, Figures~\ref{fig:main-psnr}d \& \ref{fig:main-ssim}d show PSNR and SSIM scores obtained on fastMRI knee reconstructions, while Figure~\ref{fig:main} (bottom row) shows the accompanying qualitative plots. This anatomy is challenging especially because of the poor signal-to-noise ratio conditions, which can be seen even in the ground-truth image. It can be noticed that this is the most severe shift for all methods, but our approach still shows the best performance at $R={2,4}$ and a significantly lower variance.
Appendix~\ref{app:knees} shows more examples of knee reconstructions with and without fat suppression, and Figure~\ref{fig:fs-ssim-psnr} shows metrics on fat suppressed knees. 

\subsection{Uncertainty Estimation}\label{sec:uncertainty}
Our method can also provide uncertainty estimates for each reconstructed pixel by running multiple reconstruction samplers. For a given observation $y$, we can obtain independent samples $\xhat_1, \cdots, \xhat_K \sim \mu(\cdot | y)$, for $K$ sufficiently large. Now, using the conditional mean estimate $\Bar{x} = \sum_{i=1}^K \xhat_i / K$, we can compute the pixel-wise standard deviation $\sqrt{\sum_{i=1}^K | \xhat_i - \Bar{x} |^2 / K}$, and this gives an estimate of the error in each pixel. As shown in Fig~\ref{fig:tear}, the pixel-wise standard deviation is a good estimate of the ground truth error $|x^* - \Bar{x}|$. Additionally, notice that the reconstructions are able to recover fine details such as the annotated meniscus tear\footnote{\url{https://discuss.fastmri.org/t/219}} in Fig~\ref{fig:tear} and predict low uncertainty for these features. 

Figure~\ref{fig:uncertainty-app} in Appendix~\ref{app:knees} shows another example of an annotated meniscus tear. Figures~\ref{fig:tear-baslines-1} and~\ref{fig:tear-baslines-3} show comparisons with baselines on the same examples.

\subsection{Radiologist Study}\label{sec:radiologist}
We have conducted a preliminary blind assessment of overall image quality with two board-certified radiologists and one faculty member who uses neuroimaging for their research. These experts were \emph{not} involved in our research. We have found that our algorithm was ranked best for knee scans, and tied with the baselines for abdominal and brain scans, supporting our robustness claims in the paper. For more details, please see Appendix~\ref{app:radiologist}.

\section{Limitations}\label{sec:limitations}
We reported PSNR and SSIM values as they are correlated with radiologist evaluation upto an extent, and our preliminary radiologist study in Section~\ref{sec:radiologist} suggests the feasibility of clinical adoption.
These metrics do not capture the needs of real-world radiologists, and a more detailed study is required before the proposed techniques can be clinically adopted. 

Though promising, our initial results were still limited to fast spin-echo imaging only and all data were retrospectively under-sampled. Further study is required to demonstrate prospective performance in a larger body of heterogeneous MRI data. Our method also currently requires a high compute cost at inference time, as well as the need for a pre-trained generative model. Clinical use requires fast reconstruction in addition to fast scanning. Future work should investigate whether score-based models can be trained without a fully-sampled training set as well as investigate approaches to reducing computation time.

Finally, there are potential issues related to discrimination. Specifically, it is possible that the quality of the reconstructed images varies across protected attributes, such as gender or race~\cite{larrazabal2020gender}.

\section{Conclusions}
This paper reports the first successful application of the CSGM framework for robust multi-coil MR image reconstruction under realistic sampling conditions, and provides theoretical evidence for the robustness of posterior sampling. Our score-based model was trained on a small subset of brain MRI scans without any explicit information about the sampling scheme. This shows state-of-the-art performance under severe distributional shifts, making our model applicable in a wide range of clinical settings. 

Our method shows a considerable degree of generalization to out-of-distribution samples such as abdomen and knee MRI, even when trained exclusively on brain MRI. Notably, these scans were acquired using different MRI vendors with different pulse sequence parameters and at different institutions. We postulate that adding a small set of diverse training samples to our generative model could further improve robustness, and we hypothesize that these samples may not necessarily be restricted to MR images.

The results presented in this work represent an important step to applying deep learning models in the clinic, as there is a natural variation in sampling, image orientation, receive coils, scanner hardware, and anatomy in clinical practice. 

\section{Acknowledgements} 
Ajil Jalal, Giannis Daras and Alex Dimakis have been supported by NSF Grants CCF 1763702, 1934932, AF 1901281, 2008710, 2019844, the NSF IFML 2019844 award as well as research gifts by Western Digital, Interdigital, WNCG and MLL, computing resources from TACC and the Archie Straiton Fellowship.

Eric Price has been supported by NSF Award CCF-1751040 (CAREER), NSF Award CCF-2008868, and NSF IFML 2019844.

Marius Arvinte and Jon Tamir have been supported by NSF IFML 2019844 award, ONR grant N00014-19-1-2590, NIH Grant U24EB029240, and an AWS Machine Learning Research Award.

We thank the anonymous NeurIPS reviewers for their helpful and considerate feedback.

Finally, we would like to thank the experts who graciously helped with our image assessment study.

\bibliography{main}

\begin{thebibliography}{10}

\bibitem{mridataorg}
\url{http://mridata.org/}.

\bibitem{pingouin-stats}
\url{https://pingouin-stats.org/generated/pingouin.intraclass_corr.html}.

\bibitem{aggarwal2018modl}
Hemant~K Aggarwal, Merry~P Mani, and Mathews Jacob.
\newblock Modl: Model-based deep learning architecture for inverse problems.
\newblock {\em IEEE transactions on medical imaging}, 38(2):394--405, 2018.

\bibitem{antun2020instabilities}
Vegard Antun, Francesco Renna, Clarice Poon, Ben Adcock, and Anders~C Hansen.
\newblock On instabilities of deep learning in image reconstruction and the
  potential costs of ai.
\newblock {\em Proceedings of the National Academy of Sciences},
  117(48):30088--30095, 2020.

\bibitem{arjovsky2017wasserstein}
Martin Arjovsky, Soumith Chintala, and L{\'e}on Bottou.
\newblock Wasserstein gan.
\newblock {\em arXiv preprint arXiv:1701.07875}, 2017.

\bibitem{asim2019invertible}
Muhammad Asim, Ali Ahmed, and Paul Hand.
\newblock Invertible generative models for inverse problems: mitigating
  representation error and dataset bias.
\newblock {\em arXiv preprint arXiv:1905.11672}, 2019.

\bibitem{asim2018solving}
Muhammad Asim, Fahad Shamshad, and Ali Ahmed.
\newblock Solving bilinear inverse problems using deep generative priors.
\newblock {\em CoRR, abs/1802.04073}, 3(4):8, 2018.

\bibitem{bakry1985diffusions}
Dominique Bakry and Michel {\'E}mery.
\newblock Diffusions hypercontractives.
\newblock In {\em Seminaire de probabilit{\'e}s XIX 1983/84}, pages 177--206.
  Springer, 1985.

\bibitem{balevi2020high}
Eren Balevi, Akash Doshi, Ajil Jalal, Alexandros Dimakis, and Jeffrey~G
  Andrews.
\newblock High dimensional channel estimation using deep generative networks.
\newblock {\em IEEE Journal on Selected Areas in Communications}, 39(1):18--30,
  2020.

\bibitem{baraniuk2010model}
Richard~G Baraniuk, Volkan Cevher, Marco~F Duarte, and Chinmay Hegde.
\newblock Model-based compressive sensing.
\newblock {\em IEEE Transactions on Information Theory}, 56(4):1982--2001,
  2010.

\bibitem{baraniuk2009random}
Richard~G Baraniuk and Michael~B Wakin.
\newblock Random projections of smooth manifolds.
\newblock {\em Foundations of computational mathematics}, 9(1):51--77, 2009.

\bibitem{bickel2009simultaneous}
Peter~J Bickel, Ya’acov Ritov, and Alexandre~B Tsybakov.
\newblock Simultaneous analysis of lasso and dantzig selector.
\newblock {\em The Annals of Statistics}, 37(4):1705--1732, 2009.

\bibitem{bora2017compressed}
Ashish Bora, Ajil Jalal, Eric Price, and Alexandros~G Dimakis.
\newblock Compressed sensing using generative models.
\newblock In {\em Proceedings of the 34th International Conference on Machine
  Learning-Volume 70}, pages 537--546. JMLR. org, 2017.

\bibitem{bora2018ambientgan}
Ashish Bora, Eric Price, and Alexandros~G Dimakis.
\newblock Ambientgan: Generative models from lossy measurements.
\newblock {\em ICLR}, 2:5, 2018.

\bibitem{candes2008restricted}
Emmanuel~J Candes.
\newblock The restricted isometry property and its implications for compressed
  sensing.
\newblock {\em Comptes rendus mathematique}, 346(9-10):589--592, 2008.

\bibitem{champion2008wasserstein}
Thierry Champion, Luigi De~Pascale, and Petri Juutinen.
\newblock The $\infty$-{W}asserstein distance: Local solutions and existence of
  optimal transport maps.
\newblock {\em SIAM Journal on Mathematical Analysis}, 40(1):1--20, 2008.

\bibitem{cole2021fast}
Elizabeth~K Cole, Frank Ong, Shreyas~S Vasanawala, and John~M Pauly.
\newblock Fast unsupervised mri reconstruction without fully-sampled ground
  truth data using generative adversarial networks.
\newblock In {\em Proceedings of the IEEE/CVF International Conference on
  Computer Vision}, pages 3988--3997, 2021.

\bibitem{daras2021intermediate}
Giannis Daras, Joseph Dean, Ajil Jalal, and Alexandros~G Dimakis.
\newblock Intermediate layer optimization for inverse problems using deep
  generative models.
\newblock {\em International Conference on Machine Learning}, 2021.

\bibitem{darestani2021measuring}
Mohammad~Zalbagi Darestani, Akshay Chaudhari, and Reinhard Heckel.
\newblock Measuring robustness in deep learning based compressive sensing.
\newblock {\em International Conference on Machine Learning}, 2021.

\bibitem{deshmane2012parallel}
Anagha Deshmane, Vikas Gulani, Mark~A Griswold, and Nicole Seiberlich.
\newblock Parallel mr imaging.
\newblock {\em Journal of Magnetic Resonance Imaging}, 36(1):55--72, 2012.

\bibitem{dhar2018modeling}
Manik Dhar, Aditya Grover, and Stefano Ermon.
\newblock Modeling sparse deviations for compressed sensing using generative
  models.
\newblock {\em arXiv preprint arXiv:1807.01442}, 2018.

\bibitem{doneva2021csmri}
Mariya Doneva.
\newblock Mathematical models for magnetic resonance imaging reconstruction: An
  overview of the approaches, problems, and future research areas.
\newblock {\em IEEE Signal Processing Magazine}, 37(1):24--32, 2020.

\bibitem{donoho2006compressed}
David~L Donoho.
\newblock Compressed sensing.
\newblock {\em IEEE Transactions on information theory}, 52(4):1289--1306,
  2006.

\bibitem{mardani2021mriuncertainty}
Vineet Edupuganti, Morteza Mardani, Shreyas Vasanawala, and John Pauly.
\newblock Uncertainty quantification in deep mri reconstruction.
\newblock {\em IEEE Transactions on Medical Imaging}, 40(1):239--250, 2021.

\bibitem{eldar2009robust}
Yonina~C Eldar and Moshe Mishali.
\newblock Robust recovery of signals from a structured union of subspaces.
\newblock {\em IEEE Transactions on Information Theory}, 55(11):5302--5316,
  2009.

\bibitem{fletcher2018plug}
Alyson~K Fletcher, Parthe Pandit, Sundeep Rangan, Subrata Sarkar, and Philip
  Schniter.
\newblock Plug-in estimation in high-dimensional linear inverse problems: A
  rigorous analysis.
\newblock In {\em Advances in Neural Information Processing Systems}, pages
  7440--7449, 2018.

\bibitem{fletcher2018inference}
Alyson~K Fletcher, Sundeep Rangan, and Philip Schniter.
\newblock Inference in deep networks in high dimensions.
\newblock In {\em 2018 IEEE International Symposium on Information Theory
  (ISIT)}, pages 1884--1888. IEEE, 2018.

\bibitem{gomez2019fast}
Fabian~Latorre G{\'o}mez, Armin Eftekhari, and Volkan Cevher.
\newblock Fast and provable admm for learning with generative priors.
\newblock {\em arXiv preprint arXiv:1907.03343}, 2019.

\bibitem{goodfellow2014generative}
Ian Goodfellow, Jean Pouget-Abadie, Mehdi Mirza, Bing Xu, David Warde-Farley,
  Sherjil Ozair, Aaron Courville, and Yoshua Bengio.
\newblock Generative adversarial nets.
\newblock In {\em Advances in neural information processing systems}, pages
  2672--2680, 2014.

\bibitem{griswold2002grappa}
Mark~A. Griswold, Peter~M. Jakob, Robin~M. Heidemann, Mathias Nittka, Vladimir
  Jellus, Jianmin Wang, Berthold Kiefer, and Axel Haase.
\newblock Generalized autocalibrating partially parallel acquisitions (grappa).
\newblock {\em Magnetic Resonance in Medicine}, 47(6):1202--1210, 2002.

\bibitem{guerquin2011realistic}
Matthieu Guerquin-Kern, Laurent Lejeune, Klaas~Paul Pruessmann, and Michael
  Unser.
\newblock Realistic analytical phantoms for parallel magnetic resonance
  imaging.
\newblock {\em IEEE Transactions on Medical Imaging}, 31(3):626--636, 2011.

\bibitem{hammernik2018learning}
Kerstin Hammernik, Teresa Klatzer, Erich Kobler, Michael~P Recht, Daniel~K
  Sodickson, Thomas Pock, and Florian Knoll.
\newblock Learning a variational network for reconstruction of accelerated mri
  data.
\newblock {\em Magnetic resonance in medicine}, 79(6):3055--3071, 2018.

\bibitem{hammernik2021systematic}
Kerstin Hammernik, Jo~Schlemper, Chen Qin, Jinming Duan, Ronald~M Summers, and
  Daniel Rueckert.
\newblock Systematic evaluation of iterative deep neural networks for fast
  parallel mri reconstruction with sensitivity-weighted coil combination.
\newblock {\em Magnetic Resonance in Medicine}, 2021.

\bibitem{hand2019global}
Paul Hand and Babhru Joshi.
\newblock Global guarantees for blind demodulation with generative priors.
\newblock In {\em Advances in Neural Information Processing Systems}, pages
  11531--11541, 2019.

\bibitem{hand2018phase}
Paul Hand, Oscar Leong, and Vlad Voroninski.
\newblock Phase retrieval under a generative prior.
\newblock In {\em Advances in Neural Information Processing Systems}, pages
  9136--9146, 2018.

\bibitem{hand2017global}
Paul Hand and Vladislav Voroninski.
\newblock Global guarantees for enforcing deep generative priors by empirical
  risk.
\newblock {\em arXiv preprint arXiv:1705.07576}, 2017.

\bibitem{heckel2018deep}
Reinhard Heckel and Paul Hand.
\newblock Deep decoder: Concise image representations from untrained
  non-convolutional networks.
\newblock {\em arXiv preprint arXiv:1810.03982}, 2018.

\bibitem{heckel2020compressive}
Reinhard Heckel and Mahdi Soltanolkotabi.
\newblock Compressive sensing with un-trained neural networks: Gradient descent
  finds the smoothest approximation.
\newblock {\em arXiv preprint arXiv:2005.03991}, 2020.

\bibitem{hegde2018algorithmic}
Chinmay Hegde.
\newblock Algorithmic aspects of inverse problems using generative models.
\newblock In {\em 2018 56th Annual Allerton Conference on Communication,
  Control, and Computing (Allerton)}, pages 166--172. IEEE, 2018.

\bibitem{hegde2008random}
Chinmay Hegde, Michael Wakin, and Richard~G Baraniuk.
\newblock Random projections for manifold learning.
\newblock In {\em Advances in neural information processing systems}, pages
  641--648, 2008.

\bibitem{huang2008software}
Feng Huang, Sathya Vijayakumar, Yu~Li, Sarah Hertel, and George~R Duensing.
\newblock A software channel compression technique for faster reconstruction
  with many channels.
\newblock {\em Magnetic resonance imaging}, 26(1):133--141, 2008.

\bibitem{hussein2020image}
Shady~Abu Hussein, Tom Tirer, and Raja Giryes.
\newblock Image-adaptive gan based reconstruction.
\newblock In {\em Proceedings of the AAAI Conference on Artificial
  Intelligence}, volume~34, pages 3121--3129, 2020.

\bibitem{iyer2020autoespirit}
Siddharth Iyer, Frank Ong, Kawin Setsompop, Mariya Doneva, and Michael Lustig.
\newblock Sure-based automatic parameter selection for espirit calibration.
\newblock {\em Magnetic Resonance in Medicine}, 84(6):3423--3437, 2020.

\bibitem{doneva2020ieeespsguest}
Mathews Jacob, Jong~Chul Ye, Leslie Ying, and Mariya Doneva.
\newblock Computational mri: Compressive sensing and beyond [from the guest
  editors].
\newblock {\em IEEE Signal Processing Magazine}, 37(1):21--23, 2020.

\bibitem{jalal2021instance}
Ajil Jalal, Sushrut Karmalkar, Alexandros~G Dimakis, and Eric Price.
\newblock Instance-optimal compressed sensing via posterior sampling.
\newblock {\em International Conference on Machine Learning}, 2021.

\bibitem{jalal2021fairness}
Ajil Jalal, Sushrut Karmalkar, Jessica Hoffmann, Alexandros~G Dimakis, and Eric
  Price.
\newblock Fairness for image generation with uncertain sensitive attributes.
\newblock {\em International Conference on Machine Learning}, 2021.

\bibitem{jalal2020robust}
Ajil Jalal, Liu Liu, Alexandros~G Dimakis, and Constantine Caramanis.
\newblock Robust compressed sensing using generative models.
\newblock {\em Advances in Neural Information Processing Systems}, 33, 2020.

\bibitem{jalali2019solving}
Shirin Jalali and Xin Yuan.
\newblock Solving linear inverse problems using generative models.
\newblock In {\em 2019 IEEE International Symposium on Information Theory
  (ISIT)}, pages 512--516. IEEE, 2019.

\bibitem{unser2017deepinverse}
Kyong~Hwan Jin, Michael~T. McCann, Emmanuel Froustey, and Michael Unser.
\newblock Deep convolutional neural network for inverse problems in imaging.
\newblock {\em IEEE Transactions on Image Processing}, 26(9):4509--4522, 2017.

\bibitem{kabkab2018task}
Maya Kabkab, Pouya Samangouei, and Rama Chellappa.
\newblock Task-aware compressed sensing with generative adversarial networks.
\newblock In {\em Thirty-Second AAAI Conference on Artificial Intelligence},
  2018.

\bibitem{kamath2019lower}
Akshay Kamath, Sushrut Karmalkar, and Eric Price.
\newblock Lower bounds for compressed sensing with generative models.
\newblock {\em arXiv preprint arXiv:1912.02938}, 2019.

\bibitem{karras2019style}
Tero Karras, Samuli Laine, and Timo Aila.
\newblock A style-based generator architecture for generative adversarial
  networks.
\newblock In {\em Proceedings of the IEEE/CVF Conference on Computer Vision and
  Pattern Recognition}, pages 4401--4410, 2019.

\bibitem{kelkar2021prior}
Varun~A Kelkar and Mark~A Anastasio.
\newblock Prior image-constrained reconstruction using style-based generative
  models.
\newblock {\em arXiv preprint arXiv:2102.12525}, 2021.

\bibitem{kelkar2021compressible}
Varun~A Kelkar, Sayantan Bhadra, and Mark~A Anastasio.
\newblock Compressible latent-space invertible networks for generative
  model-constrained image reconstruction.
\newblock {\em IEEE Transactions on Computational Imaging}, 7:209--223, 2021.

\bibitem{kingma2013auto}
Diederik~P Kingma and Max Welling.
\newblock Auto-encoding variational bayes.
\newblock {\em arXiv preprint arXiv:1312.6114}, 2013.

\bibitem{knoll2020fastmri}
Florian Knoll, Jure Zbontar, Anuroop Sriram, Matthew~J Muckley, Mary Bruno,
  Aaron Defazio, Marc Parente, Krzysztof~J Geras, Joe Katsnelson, Hersh
  Chandarana, et~al.
\newblock fastmri: A publicly available raw k-space and dicom dataset of knee
  images for accelerated mr image reconstruction using machine learning.
\newblock {\em Radiology: Artificial Intelligence}, 2(1):e190007, 2020.

\bibitem{larrazabal2020gender}
Agostina~J Larrazabal, Nicol{\'a}s Nieto, Victoria Peterson, Diego~H Milone,
  and Enzo Ferrante.
\newblock Gender imbalance in medical imaging datasets produces biased
  classifiers for computer-aided diagnosis.
\newblock {\em Proceedings of the National Academy of Sciences},
  117(23):12592--12594, 2020.

\bibitem{lei2019inverting}
Qi~Lei, Ajil Jalal, Inderjit~S Dhillon, and Alexandros~G Dimakis.
\newblock Inverting deep generative models, one layer at a time.
\newblock In {\em Advances in Neural Information Processing Systems}, pages
  13910--13919, 2019.

\bibitem{lin2017refinenet}
Guosheng Lin, Anton Milan, Chunhua Shen, and Ian Reid.
\newblock Refinenet: Multi-path refinement networks for high-resolution
  semantic segmentation.
\newblock In {\em Proceedings of the IEEE conference on computer vision and
  pattern recognition}, pages 1925--1934, 2017.

\bibitem{liu2020sample}
Zhaoqiang Liu, Selwyn Gomes, Avtansh Tiwari, and Jonathan Scarlett.
\newblock Sample complexity bounds for 1-bit compressive sensing and binary
  stable embeddings with generative priors.
\newblock {\em arXiv preprint arXiv:2002.01697}, 2020.

\bibitem{liu2019information}
Zhaoqiang Liu and Jonathan Scarlett.
\newblock Information-theoretic lower bounds for compressive sensing with
  generative models.
\newblock {\em arXiv preprint arXiv:1908.10744}, 2019.

\bibitem{lustig2007sparse}
Michael Lustig, David Donoho, and John~M Pauly.
\newblock Sparse mri: The application of compressed sensing for rapid mr
  imaging.
\newblock {\em Magnetic Resonance in Medicine: An Official Journal of the
  International Society for Magnetic Resonance in Medicine}, 58(6):1182--1195,
  2007.

\bibitem{mardani2017deep}
Morteza Mardani, Enhao Gong, Joseph~Y Cheng, Shreyas Vasanawala, Greg
  Zaharchuk, Marcus Alley, Neil Thakur, Song Han, William Dally, John~M Pauly,
  et~al.
\newblock Deep generative adversarial networks for compressed sensing automates
  mri.
\newblock {\em arXiv preprint arXiv:1706.00051}, 2017.

\bibitem{mardani2018deep}
Morteza Mardani, Enhao Gong, Joseph~Y Cheng, Shreyas~S Vasanawala, Greg
  Zaharchuk, Lei Xing, and John~M Pauly.
\newblock Deep generative adversarial neural networks for compressive sensing
  mri.
\newblock {\em IEEE transactions on medical imaging}, 38(1):167--179, 2018.

\bibitem{mason2019comparison}
Allister Mason, James Rioux, Sharon~E Clarke, Andreu Costa, Matthias Schmidt,
  Valerie Keough, Thien Huynh, and Steven Beyea.
\newblock Comparison of objective image quality metrics to expert
  radiologists’ scoring of diagnostic quality of mr images.
\newblock {\em IEEE transactions on medical imaging}, 39(4):1064--1072, 2019.

\bibitem{muckleyfastmri2020}
Matthew~J. Muckley, Bruno Riemenschneider, Alireza Radmanesh, Sunwoo Kim, Geunu
  Jeong, Jingyu Ko, Yohan Jun, Hyungseob Shin, Dosik Hwang, Mahmoud Mostapha,
  Simon Arberet, Dominik Nickel, Zaccharie Ramzi, Philippe Ciuciu, Jean-Luc
  Starck, Jonas Teuwen, Dimitrios Karkalousos, Chaoping Zhang, Anuroop Sriram,
  Zhengnan Huang, Nafissa Yakubova, Yvonne~W. Lui, and Florian Knoll.
\newblock Results of the 2020 fastmri challenge for machine learning mr image
  reconstruction.
\newblock {\em IEEE Transactions on Medical Imaging}, pages 1--1, 2021.

\bibitem{narnhofer2019inverse}
Dominik Narnhofer, Kerstin Hammernik, Florian Knoll, and Thomas Pock.
\newblock Inverse gans for accelerated mri reconstruction.
\newblock In {\em Wavelets and Sparsity XVIII}, volume 11138, page 111381A.
  International Society for Optics and Photonics, 2019.

\bibitem{ongie2020deep}
Gregory Ongie, Ajil Jalal, Christopher~A Metzler, Richard~G Baraniuk,
  Alexandros~G Dimakis, and Rebecca Willett.
\newblock Deep learning techniques for inverse problems in imaging.
\newblock {\em arXiv preprint arXiv:2005.06001}, 2020.

\bibitem{pandit2019inference}
Parthe Pandit, Mojtaba Sahraee-Ardakan, Sundeep Rangan, Philip Schniter, and
  Alyson~K Fletcher.
\newblock Inference with deep generative priors in high dimensions.
\newblock {\em arXiv preprint arXiv:1911.03409}, 2019.

\bibitem{pruessmann1999sense}
Klaas~P Pruessmann, Markus Weiger, Markus~B Scheidegger, and Peter Boesiger.
\newblock Sense: sensitivity encoding for fast mri.
\newblock {\em Magnetic Resonance in Medicine: An Official Journal of the
  International Society for Magnetic Resonance in Medicine}, 42(5):952--962,
  1999.

\bibitem{qiu2019robust}
Shuang Qiu, Xiaohan Wei, and Zhuoran Yang.
\newblock Robust one-bit recovery via relu generative networks: Improved
  statistical rates and global landscape analysis.
\newblock {\em arXiv preprint arXiv:1908.05368}, 2019.

\bibitem{bresler2011dictionarylearning}
Saiprasad Ravishankar and Yoram Bresler.
\newblock Mr image reconstruction from highly undersampled k-space data by
  dictionary learning.
\newblock {\em IEEE Transactions on Medical Imaging}, 30(5):1028--1041, 2011.

\bibitem{ravishankar2017datadriven}
Saiprasad Ravishankar and Jeffrey~A. Fessler.
\newblock Data-driven models and approaches for imaging.
\newblock In {\em Imaging and Applied Optics 2017 (3D, AIO, COSI, IS, MATH,
  pcAOP)}, page MW2C.4. Optical Society of America, 2017.

\bibitem{rick2017one}
JH~Rick~Chang, Chun-Liang Li, Barnabas Poczos, BVK Vijaya~Kumar, and Aswin~C
  Sankaranarayanan.
\newblock One network to solve them all--solving linear inverse problems using
  deep projection models.
\newblock In {\em Proceedings of the IEEE International Conference on Computer
  Vision}, pages 5888--5897, 2017.

\bibitem{rosenzweig2018simultaneous}
Sebastian Rosenzweig, Hans Christian~Martin Holme, Robin~N Wilke, Dirk Voit,
  Jens Frahm, and Martin Uecker.
\newblock Simultaneous multi-slice mri using cartesian and radial flash and
  regularized nonlinear inversion: Sms-nlinv.
\newblock {\em Magnetic resonance in medicine}, 79(4):2057--2066, 2018.

\bibitem{schlemper2017deep}
Jo~Schlemper, Jose Caballero, Joseph~V Hajnal, Anthony~N Price, and Daniel
  Rueckert.
\newblock A deep cascade of convolutional neural networks for dynamic mr image
  reconstruction.
\newblock {\em IEEE transactions on Medical Imaging}, 37(2):491--503, 2017.

\bibitem{shimron2021subtle}
Efrat Shimron, Jonathan~I Tamir, Ke~Wang, and Michael Lustig.
\newblock Subtle inverse crimes: Na$\backslash$" ively training machine
  learning algorithms could lead to overly-optimistic results.
\newblock {\em arXiv preprint arXiv:2109.08237}, 2021.

\bibitem{sodickson1997simultaneous}
Daniel~K Sodickson and Warren~J Manning.
\newblock Simultaneous acquisition of spatial harmonics (smash): fast imaging
  with radiofrequency coil arrays.
\newblock {\em Magnetic resonance in medicine}, 38(4):591--603, 1997.

\bibitem{song2019generative}
Yang Song and Stefano Ermon.
\newblock Generative modeling by estimating gradients of the data distribution.
\newblock In {\em Advances in Neural Information Processing Systems}, pages
  11918--11930, 2019.

\bibitem{song2020improved}
Yang Song and Stefano Ermon.
\newblock Improved techniques for training score-based generative models.
\newblock {\em arXiv preprint arXiv:2006.09011}, 2020.

\bibitem{song2021scorebased}
Yang Song, Jascha Sohl-Dickstein, Diederik~P Kingma, Abhishek Kumar, Stefano
  Ermon, and Ben Poole.
\newblock Score-based generative modeling through stochastic differential
  equations.
\newblock In {\em International Conference on Learning Representations}, 2021.

\bibitem{sriram2020end}
Anuroop Sriram, Jure Zbontar, Tullie Murrell, Aaron Defazio, C~Lawrence
  Zitnick, Nafissa Yakubova, Florian Knoll, and Patricia Johnson.
\newblock End-to-end variational networks for accelerated mri reconstruction.
\newblock In {\em International Conference on Medical Image Computing and
  Computer-Assisted Intervention}, pages 64--73. Springer, 2020.

\bibitem{Sriram_2020_CVPR}
Anuroop Sriram, Jure Zbontar, Tullie Murrell, C.~Lawrence Zitnick, Aaron
  Defazio, and Daniel~K. Sodickson.
\newblock Grappanet: Combining parallel imaging with deep learning for
  multi-coil mri reconstruction.
\newblock {\em Proceedings of the IEEE/CVF Conference on Computer Vision and
  Pattern Recognition (CVPR)}, June 2020.

\bibitem{sriram2020grappanet}
Anuroop Sriram, Jure Zbontar, Tullie Murrell, C~Lawrence Zitnick, Aaron
  Defazio, and Daniel~K Sodickson.
\newblock Grappanet: Combining parallel imaging with deep learning for
  multi-coil mri reconstruction.
\newblock In {\em Proceedings of the IEEE/CVF Conference on Computer Vision and
  Pattern Recognition}, pages 14315--14322, 2020.

\bibitem{tibshirani1996regression}
Robert Tibshirani.
\newblock Regression shrinkage and selection via the lasso.
\newblock {\em Journal of the Royal Statistical Society. Series B
  (Methodological)}, pages 267--288, 1996.

\bibitem{bart}
Martin Uecker, Christian Holme, Moritz Blumenthal, Xiaoqing Wang, Zhengguo Tan,
  Nick Scholand, Siddharth Iyer, Jon Tamir, and Michael Lustig.
\newblock mrirecon/bart: version 0.7.00, March 2021.

\bibitem{uecker2014espirit}
Martin Uecker, Peng Lai, Mark~J Murphy, Patrick Virtue, Michael Elad, John~M
  Pauly, Shreyas~S Vasanawala, and Michael Lustig.
\newblock Espirit—an eigenvalue approach to autocalibrating parallel mri:
  where sense meets grappa.
\newblock {\em Magnetic resonance in medicine}, 71(3):990--1001, 2014.

\bibitem{uecker2015berkeley}
Martin Uecker, Frank Ong, Jonathan~I Tamir, Dara Bahri, Patrick Virtue,
  Joseph~Y Cheng, Tao Zhang, and Michael Lustig.
\newblock Berkeley advanced reconstruction toolbox.
\newblock In {\em Proc. Intl. Soc. Mag. Reson. Med}, volume~23, 2015.

\bibitem{ulyanov2018deep}
Dmitry Ulyanov, Andrea Vedaldi, and Victor Lempitsky.
\newblock Deep image prior.
\newblock In {\em Proceedings of the IEEE conference on computer vision and
  pattern recognition}, pages 9446--9454, 2018.

\bibitem{vasanawala2010csmri}
Shreyas~S. Vasanawala, Marcus~T. Alley, Brian~A. Hargreaves, Richard~A. Barth,
  John~M. Pauly, and Michael Lustig.
\newblock Improved pediatric mr imaging with compressed sensing.
\newblock {\em Radiology}, 256(2):607--616, 2010.
\newblock PMID: 20529991.

\bibitem{villani2008optimal}
C{\'e}dric Villani.
\newblock {\em Optimal transport: old and new}, volume 338.
\newblock Springer Science \& Business Media, 2008.

\bibitem{wang2006modern}
Zhou Wang and Alan~C Bovik.
\newblock Modern image quality assessment.
\newblock {\em Synthesis Lectures on Image, Video, and Multimedia Processing},
  2(1):1--156, 2006.

\bibitem{bresler2020ieeesps}
Bihan Wen, Saiprasad Ravishankar, Luke Pfister, and Yoram Bresler.
\newblock Transform learning for magnetic resonance image reconstruction: From
  model-based learning to building neural networks.
\newblock {\em IEEE Signal Processing Magazine}, 37(1):41--53, 2020.

\bibitem{zbontar2018fastMRI}
Jure Zbontar, Florian Knoll, Anuroop Sriram, Tullie Murrell, Zhengnan Huang,
  Matthew~J. Muckley, Aaron Defazio, Ruben Stern, Patricia Johnson, Mary Bruno,
  Marc Parente, Krzysztof~J. Geras, Joe Katsnelson, Hersh Chandarana, Zizhao
  Zhang, Michal Drozdzal, Adriana Romero, Michael Rabbat, Pascal Vincent,
  Nafissa Yakubova, James Pinkerton, Duo Wang, Erich Owens, C.~Lawrence
  Zitnick, Michael~P. Recht, Daniel~K. Sodickson, and Yvonne~W. Lui.
\newblock {fastMRI}: An open dataset and benchmarks for accelerated {MRI}.
\newblock 2018.

\end{thebibliography}
\bibliographystyle{plain}


\appendix

\section{Appendix: Additional Metrics}\label{app:metrics}
\begin{figure}
    \centering
    \includegraphics[width=\columnwidth]{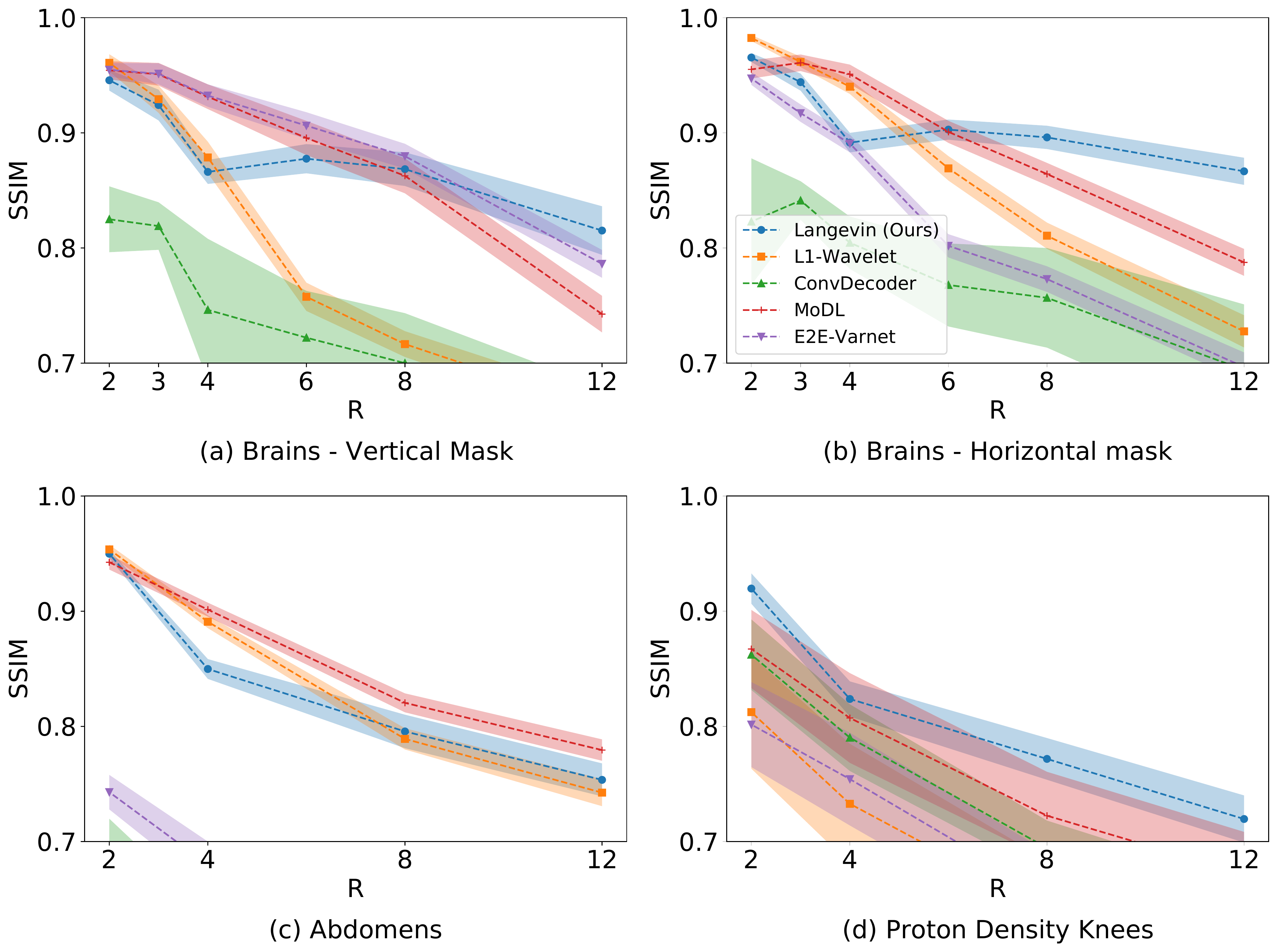}
    \caption{\small Average test SSIM in various scenarios, across a range of acceleration factors $R$. Higher $R$ indicates a smaller number of acquired measurements. Our approach mostly shows the best performance and lowest reconstruction variance both in- and out-of-distribution at test-time. Shaded regions indicate 95\% confidence intervals. Note that we trained baselines on MVUE images and hence these numerical values should not be compared with those in literature trained on RSS images (see Appendix~\ref{app:mvue-rss} for a more detailed discussion).}
    \label{fig:main-ssim}
\end{figure}

Figure~\ref{fig:main-ssim} shows the test SSIM evaluated in the same conditions as Figure~\ref{fig:main-psnr} in the main text. This highlights that our model is also robust in this metric.

\begin{figure}
    \centering
    \includegraphics[width=\columnwidth]{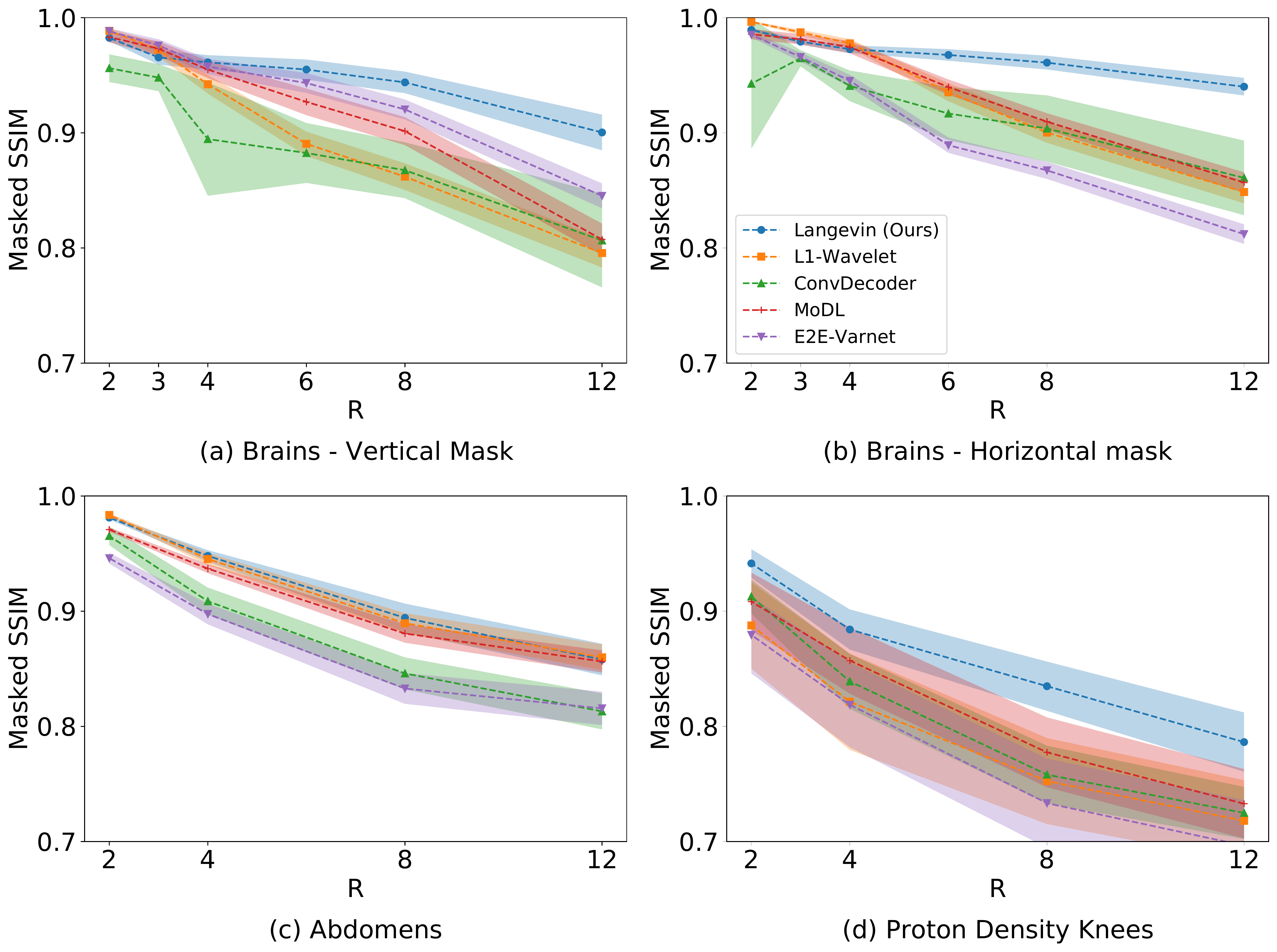}
    \caption{\small Average test SSIM, with masking, in various scenarios across a range of acceleration factors $R$. The mask zeros out all coordinates whose absolute value is smaller than 0.05 times the maximum absolute value in the fully-sampled MVUE, and this reduces the effect of noise in the background. Higher $R$ indicates a smaller number of acquired measurements. Our approach mostly shows the best performance and lowest reconstruction variance both in- and out-of-distribution at test-time. Shaded regions indicate 95\% confidence intervals. Note that we trained baselines on MVUE images and hence these numerical values should not be compared with those in literature trained on RSS images (see Appendix~\ref{app:mvue-rss} for a more detailed discussion).}
    \label{fig:main-masked-ssim}
\end{figure}

\begin{figure}
    \centering
    \includegraphics[width=\columnwidth]{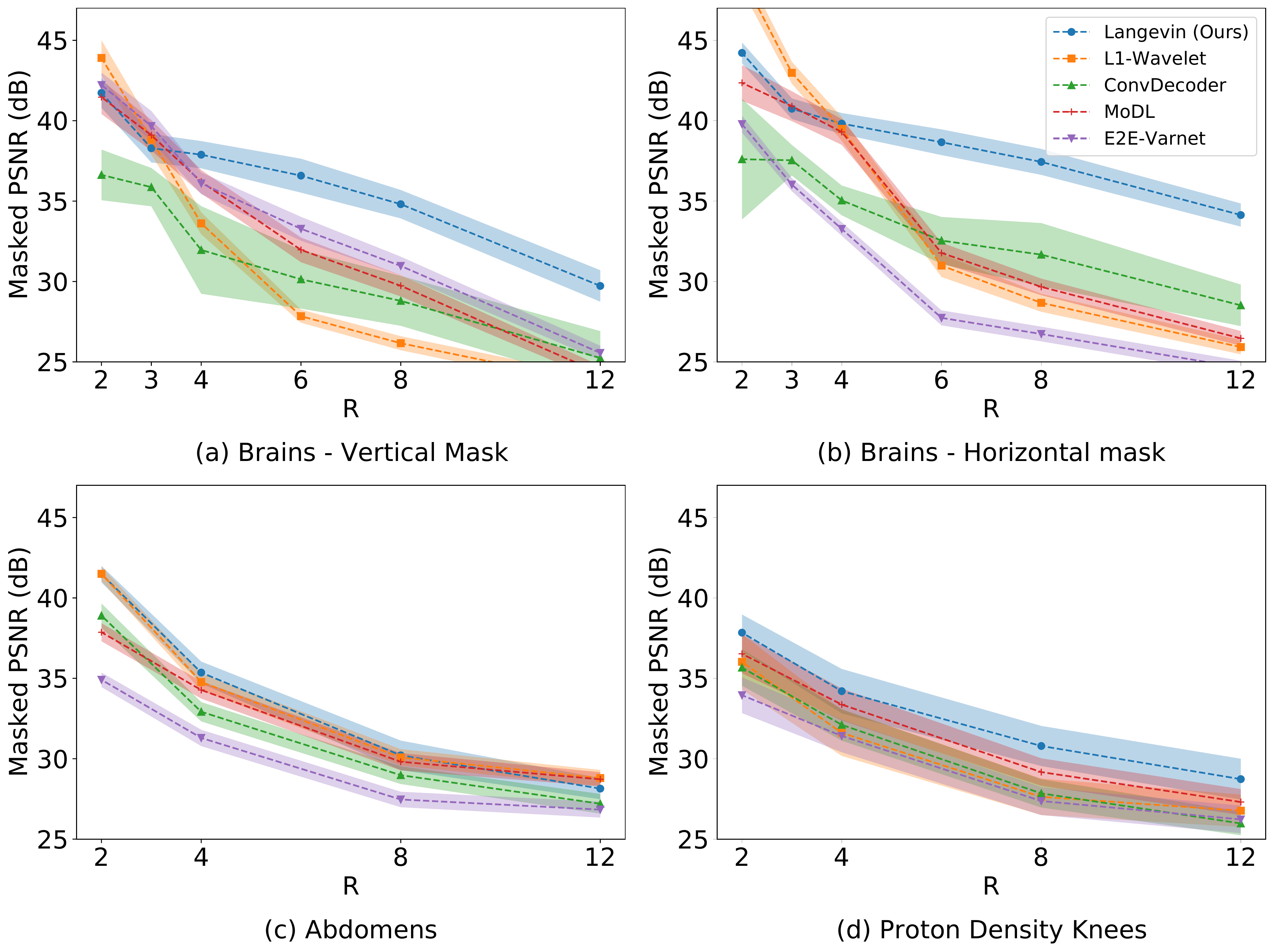}    \caption{\small Average test PSNR, with masking, in various scenarios across a range of acceleration factors $R$. The mask zeros out all coordinates whose absolute value is smaller than 0.05 times the maximum absolute value in the fully-sampled MVUE, and this reduces the effect of noise in the background. Higher $R$ indicates a smaller number of acquired measurements. Our approach mostly shows the best performance and lowest reconstruction variance both in- and out-of-distribution at test-time. Shaded regions indicate 95\% confidence intervals. Note that we trained baselines on MVUE images and hence these numerical values should not be compared with those in literature trained on RSS images (see Appendix~\ref{app:mvue-rss} for a more detailed discussion).}
    \label{fig:main-masked-psnr}
\end{figure}

We observe that our method has significant noise in the background. Hence, we also report the masked SSIM and PSNR values in Figures~\ref{fig:main-masked-ssim} and~\ref{fig:main-masked-psnr}. The mask zeros out all coordinates whose absolute value is smaller than 0.05 times the maximum absolute value in the fully-sampled MVUE.

\subsection{MVUE vs. RSS}\label{app:mvue-rss}
The difference in numerical values between our results and the publicly available fastMRI leaderboard, as well as original results in the published baseline papers baselines comes from training and evaluating all methods on MVUE instead of RSS images. This is a design choice that we have made for all baselines, since our goal is to compare with a wide range of previous methods in a fair way.

Algorithms that output a complex-valued image (such as ours and L1-Wavelet) as a solution to the optimization in Eqn~\eqref{eq:mri_recon} will artificially perform worse (w.r.t. E2E methods) when compared to the RSS ground truth, even when the output is of similar or higher quality, due to the bias in the RSS. Since there is no way to obtain a good RSS score with these algorithms, this justifies our choice to train and evaluate all methods on MVUE.

To the best of our knowledge, a rigorous, reproducible comparison between end-to-end models trained on RSS or MVUE images has not been made in prior work. The recent work of~\cite{hammernik2021systematic} has also discussed this point. To illustrate our claim of incompatibility between the two estimates, as well as the importance of qualitative inspection, we provide two simple, easy-to-verify examples.

\begin{enumerate}
    \item  We compare the fully sampled MVUE reconstruction (with ESPiRIT estimated maps) with the fully sampled RSS reconstruction, on T2 brain scans: we find that the SSIM is slightly larger than $0.8$. This is a large penalty (as per Fig.~\ref{fig:main}), even though the two images are virtually indistinguishable and known to be clinically equivalent (see discussions of SENSE vs. GRAPPA in~\cite{hammernik2021systematic}). This would unfairly penalize the family of methods that explicitly solve the inverse problem. Since E2E methods can be trained to target the MVUE directly, this justifies our choice for using the MVUE as the reference image.
    \item We point to the public knee fastMRI leaderboard at https://fastmri.org/leaderboards. Selecting "Multi-coil Knee" and "4x" acceleration, we inspect the two following submissions:
    \begin{itemize}
        \item  "zero-filling", which does zero-filling RSS reconstruction, has an SSIM of $0.804$ and considerable artifacts.
        \item  "Baseline Classical Reconstruction Model", which applies compressed sensing with the ESPiRIT algorithm, has a much poorer SSIM score of $0.6275$, but produces qualitatively superior reconstructions.
    \end{itemize}
\end{enumerate}

\section{Appendix: Theory}\label{app:theory}
\define[$\cW_q$ implies
$(\delta,\alpha)\text{-}\cW_\infty$]{lemma:winf}{Lemma}{
  If two distributions $\mu$ and $\nu$ satisfy $\cW_q(\mu, \nu) \leq \eps$ for
  some $q\geq 1$, then they satisfy
  $ (\delta,\delta)\text{-}\cW_\infty(\mu, \nu) \leq
  \eps / \delta^{1/q}$.
Futhermore, there exist distributions that satisfy 
  $ (\delta,\delta)\text{-}\cW_\infty(\mu, \nu) \leq
  \eps$, but $\cW_q(\mu, \nu) = \infty$ for all $q \geq 1$.
}
\state{lemma:winf}

\begin{proof}
Let $\Gamma$ be a coupling between $\mu, \nu$ such that $\E_{(u,v) \sim \Gamma} \left[ \norm{u - v}^q \right] \leq \eps^q$. Then an application of Markov's inequality gives
\begin{align}\label{eqn:lem1}
    \Pr[ \norm{u - v} \geq \eps / \delta^{1/q}] \leq \delta.
\end{align}

Now, we can split the distribution $\Gamma$ into two unnormalized components $\Gamma',\Gamma''$  defined as 
\begin{align*}
    \Gamma'(u,v) &= \Gamma(u,v) \bm{1}\{ \norm{u - v}< \eps / \delta^{1/q} \} ,\\
    \Gamma''(u,v) &= \Gamma(u,v) \bm{1}\{ \norm{u - v} \geq \eps / \delta^{1/q} \}.
\end{align*}

Using $\Gamma', \Gamma''$, we can define measures $\mu', \mu'', \nu', \nu''$, via
\begin{align*}
    \mu'(B) & := \Gamma'( B, \Omega),\\
    \mu''(B) & := \Gamma''( B, \Omega),\\
    \nu'(B) & := \Gamma'( \Omega, B),\\
    \nu''(B) & := \Gamma''( \Omega, B),
\end{align*}
where $B$ is any measurable set and $\Omega$ is the state-space.

Since $\Gamma$ is a valid coupling between $\mu, \nu$, and $\Gamma', \Gamma''$ are disjoint distributions, for any measurable $B\subseteq \Omega$, we have:
\begin{align*}
    \mu(B) &= \Gamma(B, \Omega) , \\
    &= \Gamma'(B, \Omega) + \Gamma''(B,\Omega),\\
    &= \mu'(B) + \mu(B''),\\
    &= \mu'(\Omega) \frac{\mu'(B)}{\mu'(\Omega)} + \mu''(\Omega) \frac{\mu''(B)}{\mu''(\Omega)}.
\end{align*}
Using Eqn~\eqref{eqn:lem1}, we can conclude that $\mu'(\Omega) \geq 1 - \delta, \mu''(\Omega) \leq \delta$.
Setting $\mu' \leftarrow \mu' / \mu'(\Omega)$ and $\mu'' \leftarrow \mu'' / \mu''(\Omega) $, we can now rewrite $\mu$ as $\mu = (1-\delta) \mu' + \delta \mu''$. A similar argument for $\nu$ gives $\nu = (1-\delta) \nu' + \delta \nu''$.

By construction, $\mu', \nu'$ can be $\cW_\infty$ coupled via $\Gamma'$ to within a distance of $\eps/ \delta^{1/q}$.
This shows that $(\delta, \delta)\text{-}\cW_\infty(\mu, \nu) \leq \eps / \delta^{1/q}$.

Now we need to show that two distributions can be close in $(\delta, \delta)\text{-}\cW_\infty$, but $\cW_q = \infty$ for all $q$. 
Consider two scalar distributions $\mu, \nu$ defined as 
\[
\mu = \begin{cases}
0 & \text{ with probability } 1 - \delta,\\
r & \text{ with probability } \delta,
\end{cases},
\]
\[
\nu = \begin{cases}
\eps & \text{ with probability } 1 - \delta,\\
-r & \text{ with probability } \delta.
\end{cases}
\]

Clearly, these distributions satisfy $(\delta, \delta)\text{-}\cW_\infty(\mu, \nu) \leq \eps$, but $\cW_q(\mu, \nu) \approx r$ for all $q$. As $r\to \infty$, we get $\cW_q (\mu, \nu) \to \infty$ for all $q\geq 1$.

\end{proof}

\subsection{Proof of Theorem~\ref{thm: main}}
In order to prove the Theorem, we make use of the following three Lemmas from~\cite{jalal2021instance}.
\define{lemma: ajil1}{Lemma}{
  \cite{jalal2021instance} For $c\in [0,1]$, let $H:= (1-c)H_0 + cH_1$ be a mixture of two
  absolutely continuous distributions $H_0,H_1$ admitting densities
  $h_0, h_1$.  Let $y$ be a sample from the distribution $H$,  such
  that  $y|z^* \sim H_{z^*}$ where $z^*\sim Bernoulli(c)$.
  
  Define $\wh{c}_{y} = \frac{ ch_1(y)}{ (1-c) h_0(y) + ch_1(y)}$, and
  let $\wh{z}|y \sim Bernoulli(\wh{c}_{y})$ be the posterior sampling of $z^*$ given $y$.
  Then we have 
  \begin{equation*}
  \Pr_{z^*, y, \wh{z}}[z^* = 0, \wh{z}=1] \leq 1 - TV(H_0, H_1).
  \end{equation*}
}
\state{lemma: ajil1}

\define{lemma: ajil2}{Lemma}{
\cite{jalal2021instance} Let $y$ be generated from $x^*$ by a Gaussian measurement process
with noise rate $\sigma$. For a fixed $\xtilde \in \R^n,$ and
parameters $\eta>0 , c \geq 4e^2$, let $P_{out}$ be a distribution
supported on the set 
\[
S_{\xtilde, out} := \{ x\in \R^n: \norm{x - \xtilde} \geq c(\eta + \sigma )  \}.
\]
Let $P_{\xtilde} $ be a distribution which is supported within an
$\eta-$radius ball centered at $\xtilde$.

For a fixed $A$, let $H_{\tilde{x}}$ denote the distribution of
$y$ when $x^* \sim P_{\tilde{x}}.$ Let $H_{out}$ denote the
corresponding distribution of $y$ when $x^* \sim P_{out}.$    
Then we have:
\begin{align*}
\E_A \left[ TV( H_{\xtilde}, H_{out})  \right] \geq 1 - 4
e^{-\frac{m}{2}\log\left( \frac{c}{4e^2} \right)} .
\end{align*}
}
\state{lemma: ajil2}
  
\define{lemma: ajil3}{Lemma}{%
\cite{jalal2021instance} Let $R,P,$ denote arbitrary distributions over $\R^n$ such
that $\cW_\infty(R,P) \leq \varepsilon$.

Let $x^* \sim R$ and $z^* \sim P$ and let $y$ and $u$ be
generated from $x^*$ and $z^*$ via a Gaussian measurement
process with $m$ measurements and noise rate $\sigma$. Let
$\xhat \sim P( \cdot | y, A)$ and $\zhat \sim P( \cdot | u, A)$.
For any $d>0$, we have
\begin{align*}
\Pr_{x^* , A , w, \xhat } \left[ \norm{x^* - \xhat} \geq d + \varepsilon \right] \leq  & e^{-\Omega(m)} + e^{\left( \frac{4\varepsilon\left(
\varepsilon + 2\sigma \right)m}{2\sigma^2} \right)}\Pr_{z^*, A
, w, \zhat }\left[ \norm{ z^* - \zhat} \geq d \right].
\end{align*}  
}
\state{lemma: ajil3}

\restate{thm: main}
\begin{proof}
	 We know from $(\delta, \alpha)\text{-}\cW_{\infty} (\mu, \nu) \leq \eps$ that there exist $\mu', \nu', \mu'', \nu''$  
	 and a finite distribution $Q$ supported on a set $S$ such that 
	\begin{enumerate}
		\item $\cW_{\infty}(\mu', \nu') \leq \eps$,
		\item $\min \{ \cW_{\infty}(\nu', Q),\cW_{\infty}(\mu', Q) \}  \leq \sigma$,
		\item $\mu = (1-\delta) \mu' + \delta \mu''$ and $\nu = (1-\alpha) \nu' + \alpha \nu''$.
	\end{enumerate}

	Suppose $\cW_{\infty}(\nu', Q) \leq \sigma$. If not, then $\cW_{\infty}(\mu', Q) \leq \sigma $, and by (1),  we see that $\cW_{\infty}(\nu', Q) \leq \sigma + \eps$, and we will use this in the proof instead. 
	By decomposing $\mu = (1- \delta) \mu' +  \delta \mu''$, we have
	\begin{align}\label{eqn: rp thm eqn 1}
	\Pr_{x^*\sim \mu , \widehat{x} \sim \nu( \cdot | y)} \left[ \norm{x^* - \widehat{x}} \geq (2c+1) \sigma + \eps \right] &
	\leq \delta + (1-\delta) \Pr_{x^*\sim \mu' ,\widehat{x} \sim \nu( \cdot | y)} \left[ \norm{x^* - \widehat{x}} \geq (2 c+1) \sigma + \eps \right].
  \end{align}

	We now bound the second term on the right hand side of the above equation.
	For this term, consider the joint distribution over $x^*, A , w, \widehat{x}$.
	By Lemma~\ref{lemma: ajil3},  we can replace $x^* \sim \mu'$ with $z^* \sim \nu'$, 
	replace $y = Ax^* + w$ with $u = A z^* + w,$ and 
	replace $\xhat \sim \nu( \cdot | A,y)$ with $\wh{z} \sim \nu( \cdot | A,u)$
	to get the following bound
  \begin{align}
    & \Pr_{x^*\sim \mu' , A , w, \widehat{x} \sim \nu( \cdot | A, y)} \left[ \norm{x^* - \widehat{x}} \geq \left( 2 c + 1 \right)\sigma + \eps \right] \leq e^{-\Omega(m)} + e^{\left( \frac{2\eps\left( \eps + 2\sigma \right)m}{\sigma^2} \right)}\Pr_{z^*\sim \nu', A , w, \widehat{z} \sim \nu(\cdot | u, A)}\left[ \norm{ z^* - \widehat{z}} \geq (2c+1)\sigma \right]\label{eqn: rp thm eqn 2}.
  \end{align}

    We now bound the second term in the right hand side of the above inequality.
    Let $\Gamma$ denote an optimal $\cW_\infty-$coupling between $\nu'$ and $Q$. 

    For each $\tilde{z} \in S$, the conditional coupling can be defined as
    \[
    \Gamma( \cdot | \tilde{z}) = \frac{\Gamma( \cdot, \tilde{z})}{Q(\tilde{z})}.
    \]
    By the $\cW_\infty$ condition, each $\Gamma( \cdot | \tilde{z})$ is supported on a ball of radius $\sigma$ around $\tilde{z}$.

    Let $ E = \{ z^*, \widehat{z} \in \R^n: \norm{ z^* - \widehat{z}} \geq \left( 2c+1 \right)\sigma\}$ denote the event that $z^*, \widehat{z}$ are far apart.
  By the coupling, we can express $\nu'$ as 
  \begin{align*}
    \nu' &= \sum_{\tilde{z} \in S} Q(\tilde{z}) \Gamma( \cdot | \tilde{z}).
  \end{align*}
  This gives 
  \begin{align*} 
    \Pr_{z^* \sim \nu', A , w, \widehat{z} \sim \nu(\cdot | A, u)} \left[ E \right] & = \sum_{\tilde{z}^* \in S} Q ( \tilde{z}^*) \E_{z^*\sim \Gamma( \cdot | \tilde{z}^*) , A , w, \widehat{z} \sim \nu(\cdot | A, u) } \left[ {1}_E \right].
  \end{align*}
  
    For each $\tilde{z}^* \in S$, we now bound $Q ( \tilde{z}^*) \E_{z^*\sim \Gamma( \cdot | \tilde{z}^*) , A , w, \widehat{z} \sim \nu(\cdot | A, u) } \left[ {1}_E \right].$ 

		For each $\tilde{z}^* \in S$, we can write $\nu$ as $\nu = \left( 1- \alpha \right) Q_{\tilde{z}^*} \nu_{\tilde{z}^*, 0} + c_{\tilde{z}^*, 1 } \nu_{\tilde{z}^*, 1} + c_{\tilde{z}^*,2} \nu_{\tilde{z}^*,2}$, where the components of the mixture are defined in the following way. The first component $\nu_{\tilde{z}^*,0}$ is $\Gamma( \cdot | \tilde{z}^*)$, the second component is supported within a $2 c \sigma$ radius of $\tilde{z}^*$, and the third component is supported outside a $2 c \sigma $ radius of $\tilde{z}^*$.

		Formally, let $B_{\tilde{z}^*}$ denote the ball of radius $c
		 \sigma$ centered at $\tilde{z}^*$, and let $B^c_{\tilde{z}^*}$
		be its complement. The constants are defined via the following
		Lebesque integrals, and the mixture components for any Borel
		measurable $B$ are defined as
		\begin{align*}
		  c_{\tilde{z}^*, 1} &:= \int_{B_{\tilde{z}^*}} d\nu - \left( 1- \alpha \right) Q_{\tilde{z}^*} \int_{B_{\tilde{z}^*}} d\Gamma( \cdot | \tilde{z}^*)  ,\\
		  \nonumber\\
		  c_{\tilde{z}^*, 2} &:= \int_{B_{\tilde{z}^*}^c} d\nu - \left( 1- \alpha \right) Q_{\tilde{z}^*} \int_{B_{\tilde{z}^*}^c} d\Gamma( \cdot | \tilde{z}^*)  ,\\
		  \nonumber\\
		  \nu_{\tilde{z}^*, 0}(B) &:= \Gamma( B \cap B_{\tilde{z}^*} | \tilde{z}^*) = \Gamma( B  | \tilde{z}^*) \text{ since } \supp(\Gamma(\cdot | \tilde{z}^*)) \subset B_{\tilde{z}^*}, \\
		  \nonumber\\
		  \nu_{\tilde{z}^*, 1}(B) &:= \begin{cases}
		    \frac{ 1}{c_{\tilde{z}^*, 1}}  \nu( B \cap B_{\tilde{z}^*})  - \frac{ 1 - \alpha }{c_{\tilde{z}^*, 1}} Q_{\tilde{z}^*} \Gamma(  B \cap B_{\tilde{z}^*} | \tilde{z}^*)  & \text{ if } c_{\tilde{z}^*, 1} > 0,\\
		    \text{do not care } & \text{ otherwise.}
		  \end{cases},\\
		  \nonumber\\
		  \nu_{\tilde{z}^*, 2}(B) &:= \begin{cases}
		    \frac{ 1}{c_{\tilde{z}^*, 2}}  \nu( B \cap B^c_{\tilde{z}^*})  - \frac{ 1 - \alpha }{c_{\tilde{z}^*, 2}} Q_{\tilde{z}^*} \Gamma(  B \cap B^c_{\tilde{z}^*} | \tilde{z}^*)  & \text{ if } c_{\tilde{z}^*, 2} > 0,\\ \text{do not care } & \text{ otherwise.}
		  \end{cases}.
		\end{align*}
		Notice that if $z^*$ is sampled from $\Gamma(\cdot | \tilde{z}^*)$, then by the $W_\infty$ condition, we have $\norm{ z^* - \tilde{z}^*} \leq \sigma$. Furthermore, if $\widehat{z}$ is $\left( 2c+1 \right)\sigma $ far from $z^*$, an application of the triangle inequality implies that it must be distributed according to $\nu_{\tilde{z}^*,2}$.
    That is, 
    \begin{align*}
			Q ( \tilde{z}^*) \E_{z^*\sim \Gamma( \cdot | \tilde{z}^*) , A , w, \widehat{z} \sim \nu(\cdot | A,u) } \left[ {1}_E \right] & \leq \E_{A , w, z^*} \Pr \left[ z^* \sim \nu_{\tilde{z}^*,0}, \widehat{z} \sim \nu_{\tilde{z}^*,2}( \cdot | u) \right] \\
			& \leq \frac{1}{1-\alpha}\E_A \left[ 1 - TV( H_{\tilde{z}^*,0}, H_{\tilde{z}^*,2}) \right],
    \end{align*}
    where $H_{\tilde{z}^*,0}, H_{\tilde{z}^*,2}$ are the push-forwards of $\nu_{\tilde{z}^*,0}, \nu_{\tilde{z}^*,2}$ for $A$ fixed and the last inequality follows from Lemma~\ref{lemma: ajil1}.

	Notice that if we sum over all $\tilde{z}^* \in S $, then the LHS
	of the above inequality is an expectation over $z^* \sim \nu'$. This
	gives:
    \begin{align*} 
      \Pr_{z^* \sim \nu', A , w, \widehat{z} \sim \nu(\cdot | u,A)} \left[ E \right] & \leq \frac{1}{1-\alpha} \sum_{\tilde{z}^* \in S} \E_{A} \left[ 1 - TV(H_{\tilde{z}^*, 0}, H_{\tilde{z}^*,2}) \right].
    \end{align*}

    Notice that $\nu_{\tilde{z}^*,0}$ is supported within an $\sigma-$ball around $\tilde{z}^*$, and $\nu_{\tilde{z}^*,2}$ is supported outside a $2c\sigma-$ball of $\tilde{z}^*$. By Lemma~\ref{lemma: ajil2} we have 
    \begin{align*}
			\E_A [TV(H_{\tilde{z}^*,0}, H_{\tilde{z}^*,2})] \geq & 1 - 4 e^{-\frac{m}{2}\log\left( \frac{c}{4e^2} \right)}.
    \end{align*}

    This implies 
    \begin{align*}
      \Pr_{z^*\sim \nu', A, w, \widehat{z} \sim \nu( \cdot | u, A )}\left[\norm{ z^* - \wh{z} } \geq (2c+1)\sigma \right] & \leq  \frac{1}{1-\alpha} \sum_{\tilde{z}^*\in S} \E_A \left[ ( 1 - TV(H_{\tilde{z}^*, 0}, H_{\tilde{z}^*,2})) \right], \\
      & \leq \frac{1}{1-\alpha}4 |S|  e^{-\frac{m}{2}\log\left( \frac{c}{4e^2} \right)},\\
			& \leq 4 e^{-\frac{m}{4}\log\left( \frac{c}{4e^2} \right)},
    \end{align*}
		where the last inequality is satisfied if $m \geq 4 \log\left(\frac{1}{1-\alpha}\right) + 4\log\left(|S|\right).$

		Substituting in Eqn~\eqref{eqn: rp thm eqn 2}, if $c > 4 \exp{\left( 2+  \frac{8\eps\left( \eps + 2\sigma \right)}{\sigma^2} \right)},$ we have
		\begin{align*}
		  \Pr_{x^*\sim \mu' , A , w, \widehat{x} \sim \nu( \cdot | A, y)} \left[ \norm{x^* - \widehat{x}} \geq \left( 2c + 1 \right)\sigma + \eps  \right] \leq & e^{-\Omega(m)}.
		\end{align*}

		This implies that there exists a set $S_{A,w}$ over $A,w$ satisfying $\Pr_{A,w}[S_{A,w}] \geq 1 - e^{-\Omega(m)},$ such that for all $A , w \in S_{A, w},$ we have
		\begin{align*}
		  \Pr_{x^*\sim \mu', \widehat{x} \sim \nu( \cdot | y)}\left[\norm{ x^* - \wh{x} } \geq (2c+1)\sigma + \eps \right] & \leq  e^{-\Omega(m)}.
		\end{align*}

		Substituting in Eqn~\eqref{eqn: rp thm eqn 1}, we have
		\begin{align*}
			\Pr_{x^*\sim \mu, \widehat{x} \sim \nu( \cdot | y)}\left[\norm{ x^* - \wh{x} } \geq (2c+1)\sigma + \eps \right] & \leq \delta +   e^{-\Omega(m)}.
		\end{align*}
		Rescaling $c$ gives us our result.
		
		At the beginning of the proof, we had assumed that $\cW_\infty(\nu',Q)\leq \sigma$.
		If instead $\cW_\infty(\mu',Q) \leq \sigma$, then we need to replace $\sigma$ in the above bound by $\sigma + \eps$.
		Rescaling $c$ in the above bound gives us the Theorem statement.

\end{proof}

\subsection{Proof of Theorem~\ref{thm:optimal}}
\restate{thm:optimal}
\begin{proof}
By the statement of the Lemma, and conditioning on the measurements $y$, we have
\begin{align*}
    1 - \delta = \Pr[ d(x^* , x') \leq \eps ] &= \E_{y } \left( \Pr[d(x^* , x') \leq \eps | y ] \right).
\end{align*}

Using a similar conditioning for the event $d(x^* , \xhat) \leq 2 \eps$, we get
\begin{align*}
    \Pr[ d(x^* , \xhat) \leq 2\eps ] &= \E_{y} \left( \Pr[ d(x^* , \xhat) \leq 2\eps | y ] \right),\\ 
   &\geq \E_{y}  \left( \Pr[d(x^* , x') \leq \eps  \land d(x' , \xhat) \leq \eps  | y ] \right),\\ 
   &= \E_{y}  \left( \Pr[ d(x^* , x') \leq \eps | y ] \cdot \Pr[ d(x' , \xhat) \leq \eps  | y ] \right),\\ 
   &= \E_{y}  \left( \Pr[ d(x^* , x') \leq \eps | y ]^2  \right),\\ 
   &\geq \left(\E_{y}  \left( \Pr[d(x^* , x') \leq \eps | y ]  \right)\right)^2,\\ 
   &= (1-\delta)^2 \geq 1 - 2\delta,
\end{align*}
where the second line follows from a triangle inequality, the third line follows since $x^*, \xhat$ are independent conditioned on $y$, the fourth line follows since $\xhat|y$ is distributed according to $x^* | y$, and the fifth line follows from Jensen's inequality.

\end{proof}

\section{Appendix: fastMRI Brain}\label{app:brains}
\subsection{Examples of Sampling Masks}
\begin{figure}[t]
    \centering
    \includegraphics[width=\columnwidth]{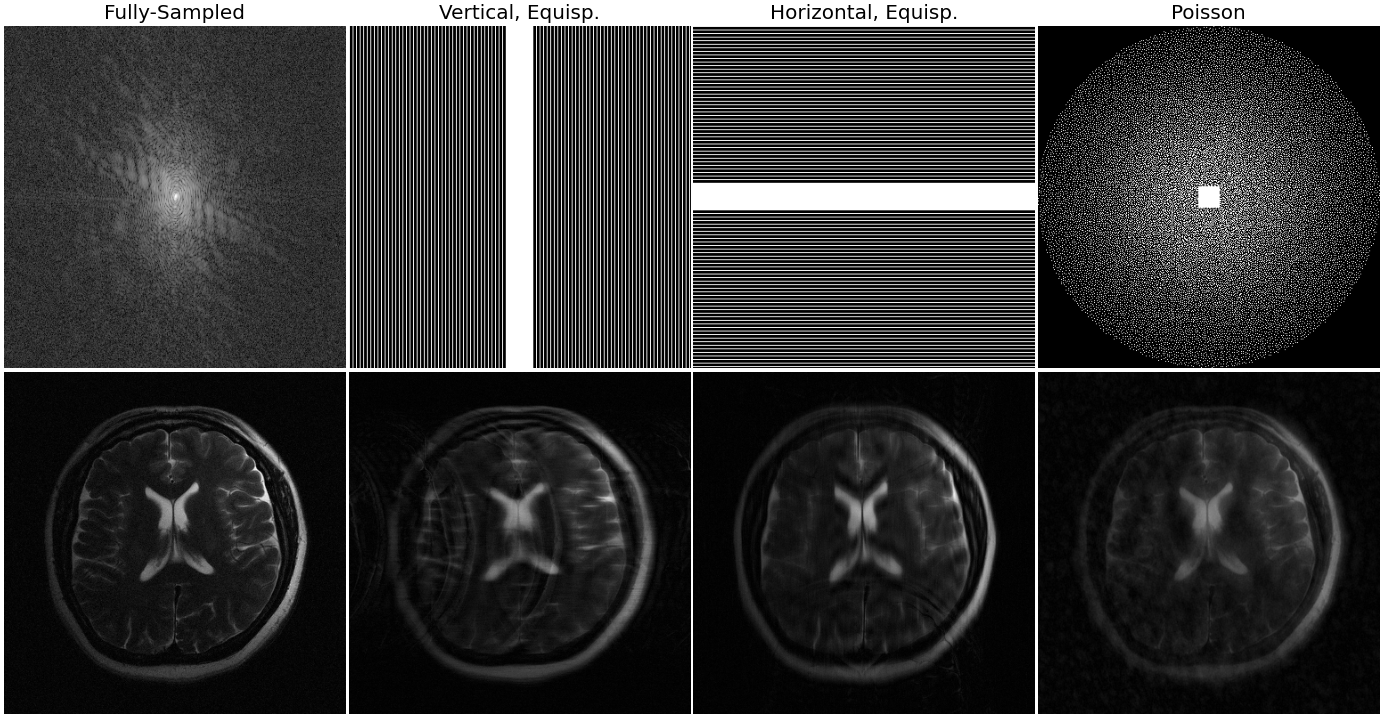}
    \caption{\small Examples of sampling patterns used throughout the experiments (top) and naive reconstructions (bottom). Top: The leftmost image shows the log-magnitude of the fully sampled k-space measurements corresponding to a single coil. The remaining images show three possible sampling masks, all with acceleration factor $R=4$ but drastically different patterns. Bottom: Each image shows the magnitude of the reconstruction obtained by a two-dimensional IFFT applied to the sampled k-space.}
    \label{fig:example_masks}
\end{figure}

Figure~\ref{fig:example_masks} shows example of some of the masks used throughout the experiments in the paper and their corresponding reconstructions. Note that the type of mask used is coupled with the scan parameters (e.g., two-dimensional slices from a three-dimensional scan will use a 2D grid of points).

We also highlight that, in all cases, a central region of the k-space is kept fully sampled and is used to estimate the coil sensitivity maps for all methods. The bottom row of Figure~\ref{fig:example_masks} shows naive reconstructions of a single coil image using the zero-filled k-space. This shows that different types of masks lead to different types of aliasing patterns in the image domain, motivating the need for robust image reconstruction algorithms.

\subsection{More Exemplar Reconstructions}
\begin{figure}
    \centering
    \includegraphics[width=\columnwidth]{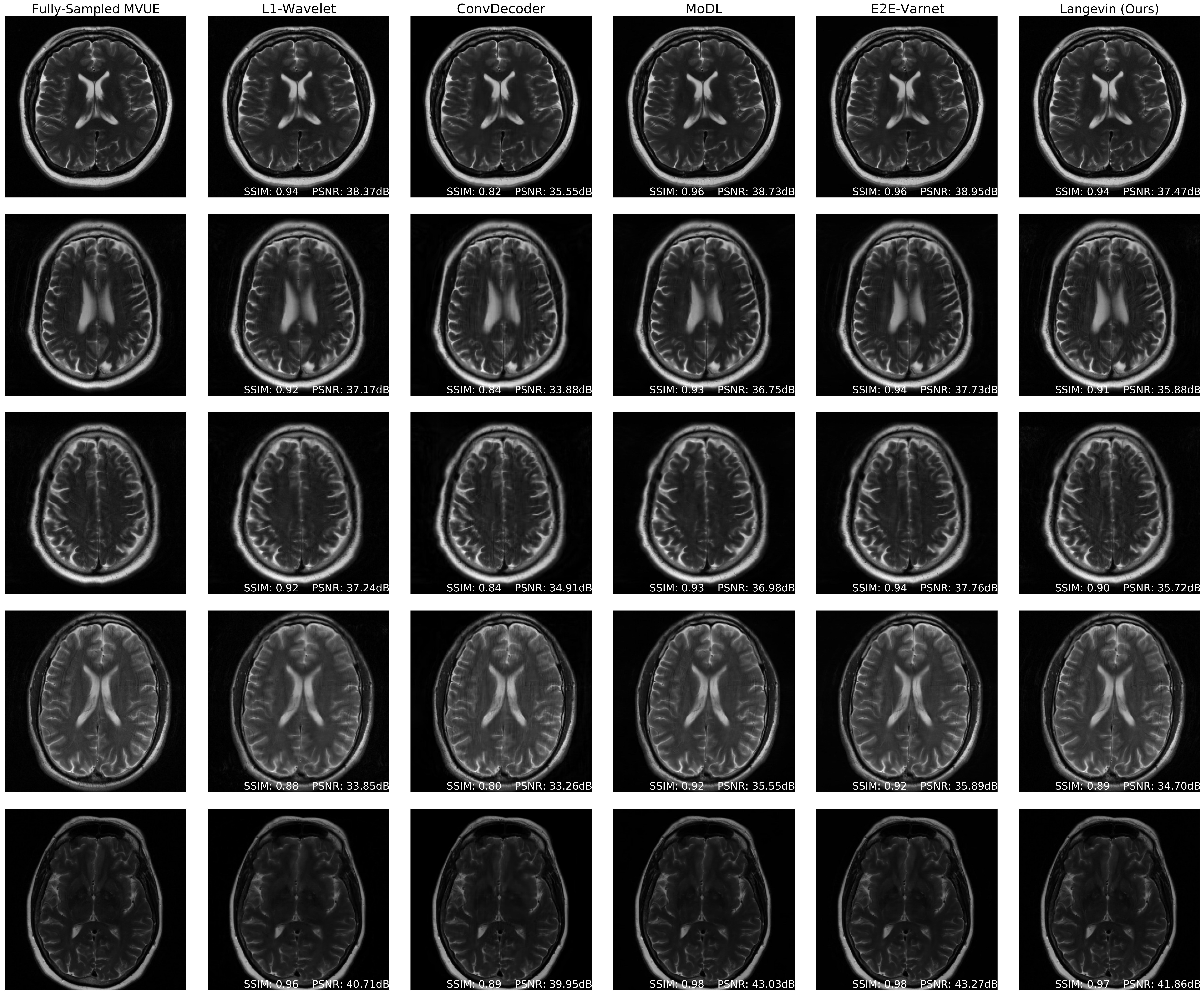}
    \caption{ In-distribution brain reconstructions, at an acceleration factor of $R=3$ and an equispaced vertical mask in k-space. Our model was trained on T2-weighted brain images from the fastMRI dataset. These results show that our method is competitive with state-of-the-art methods such as E2E-VarNet.}
    \label{fig:brain-in-3}
\end{figure}

\begin{figure}
    \centering
    \includegraphics[width=\columnwidth]{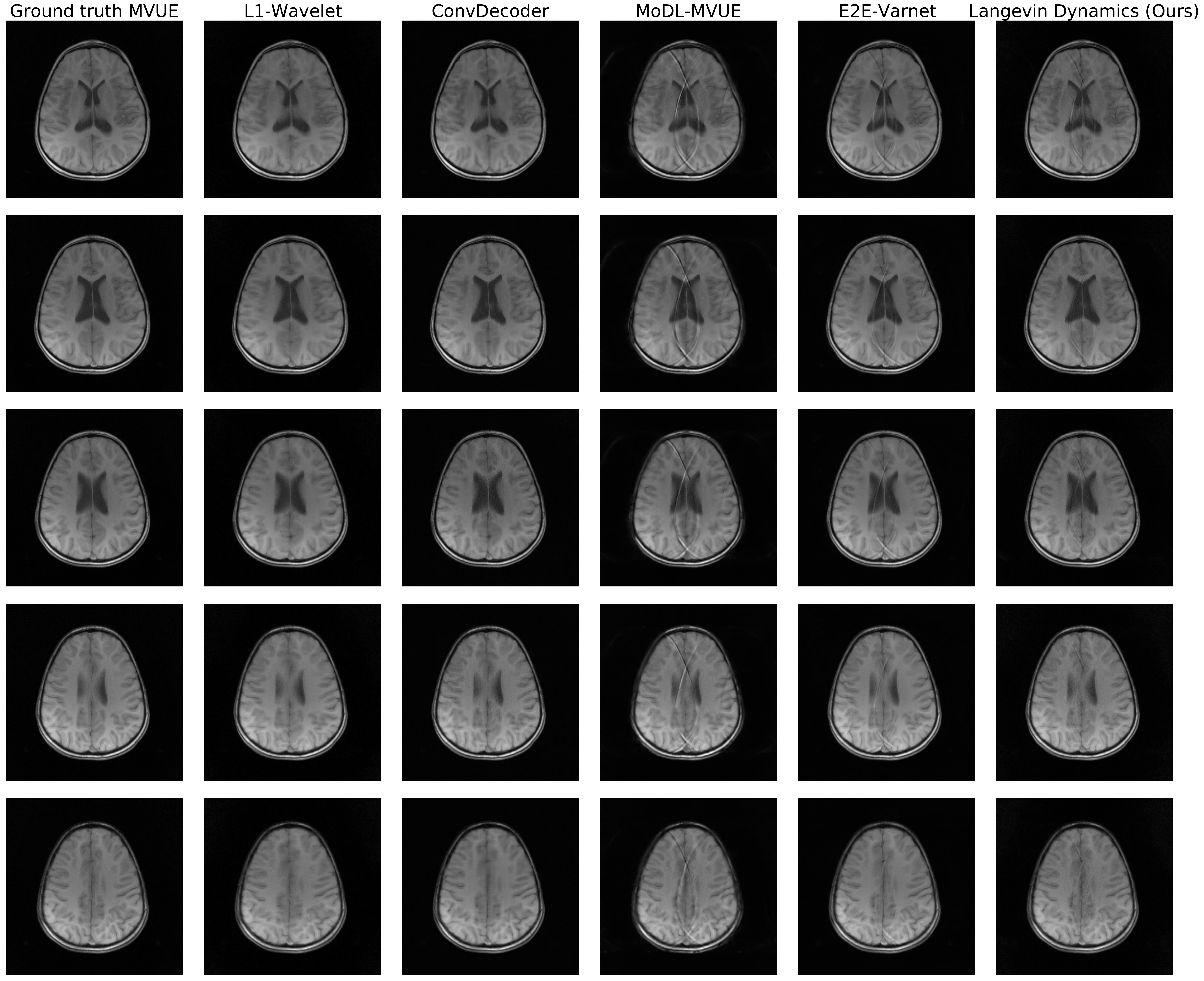}
    \caption{Brain reconstructions under a contrast shift, at an acceleration of $R=4$. Our method was trained on T2-weighted brains, while these are T1-weighted brains, and our method is clearly robust to this contrast shift.}
    \label{fig:brain-t1-4}
\end{figure}

\begin{figure}
    \centering
    \includegraphics[width=\columnwidth]{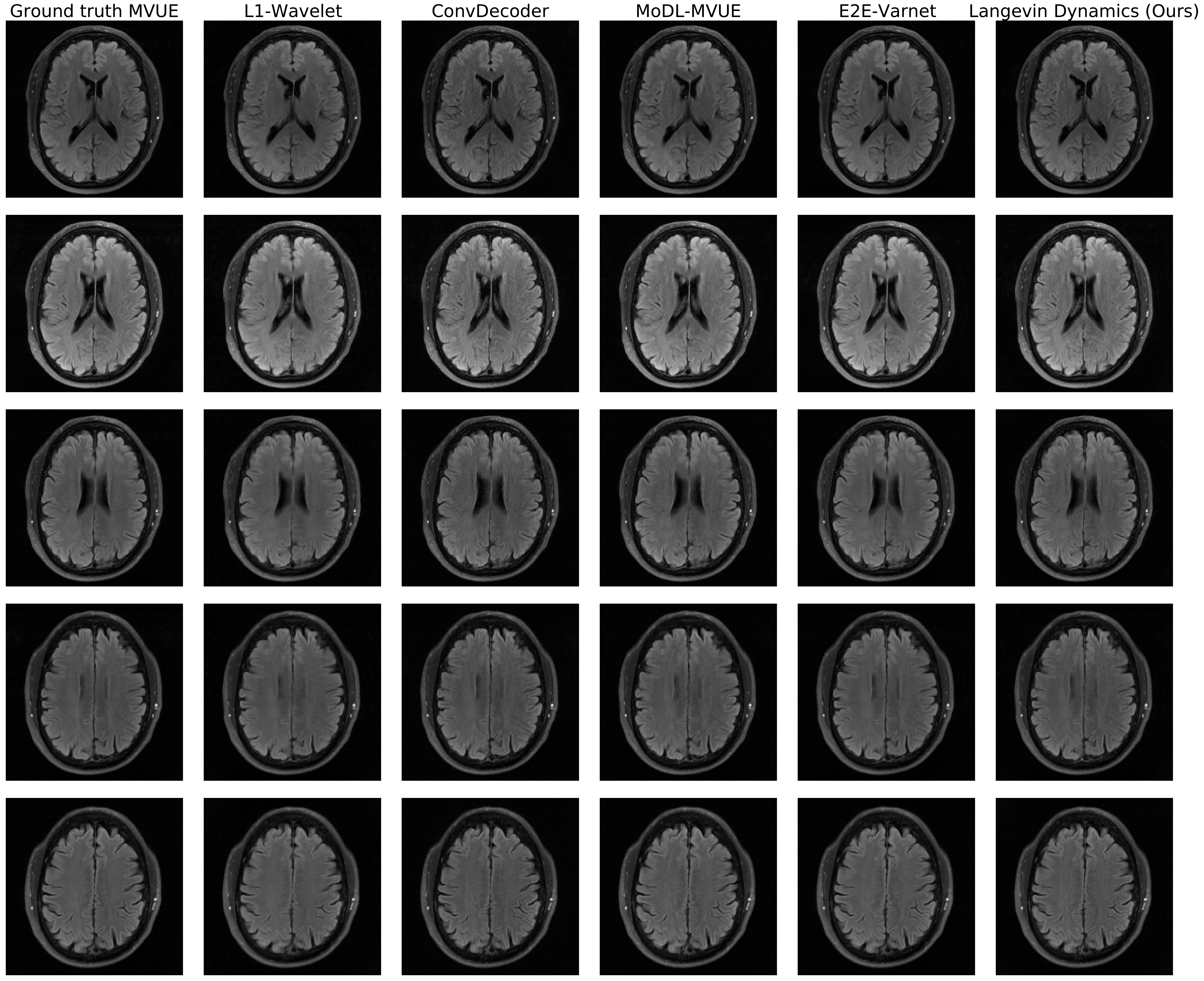}
    \caption{Brain reconstructions under a contrast shift, at an acceleration of $R=4$. Our method was trained on T2-weighted brains, while these are FLAIR brains, and our method is clearly robust to this contrast shift.}
    \label{fig:brain-flair-4}
\end{figure}

Figures~\ref{fig:brain-in-3} throughout \ref{fig:brain-flair-4} show detailed qualitative reconstructions on different brain scans from the fastMRI dataset. We highlight Figures~\ref{fig:brain-t1-4} and \ref{fig:brain-flair-4}, which represent a contrast shift from the in-distribution data (T1 and FLAIR vs. T2, respectively). Our method still produces excellent qualitative reconstructions.

\begin{figure}
    \centering
    \includegraphics[width=\columnwidth]{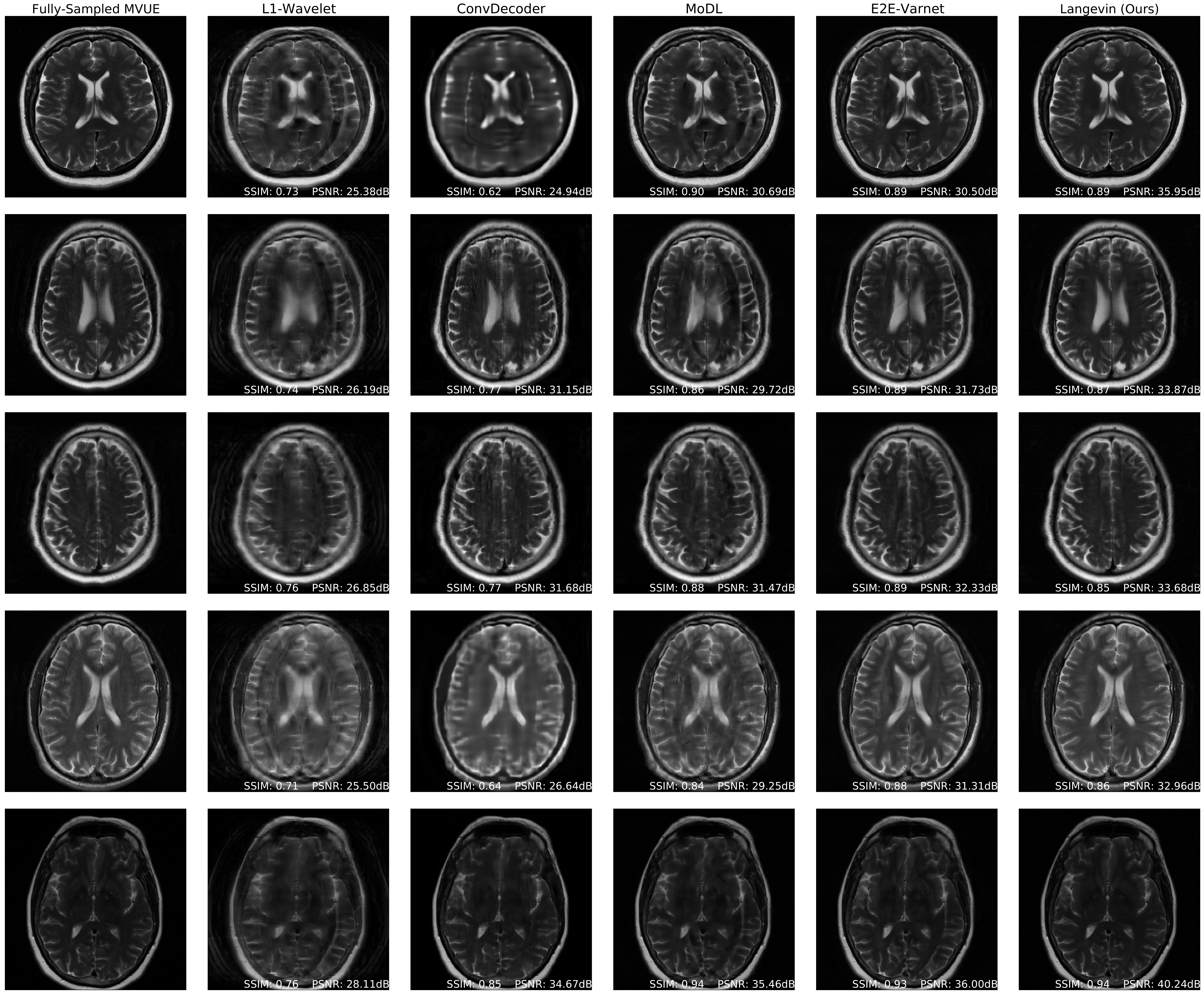}
    \caption{ In-distribution brain reconstructions, at an acceleration factor of $R=6$ and an equispaced vertical mask in k-space. Our model was trained on T2-weighted brain images from the fastMRI dataset. These results show that our method retains its performance at higher acceleration factors.}
    \label{fig:brain-in-6}
\end{figure}

\begin{figure}
    \centering
    \includegraphics[width=\columnwidth]{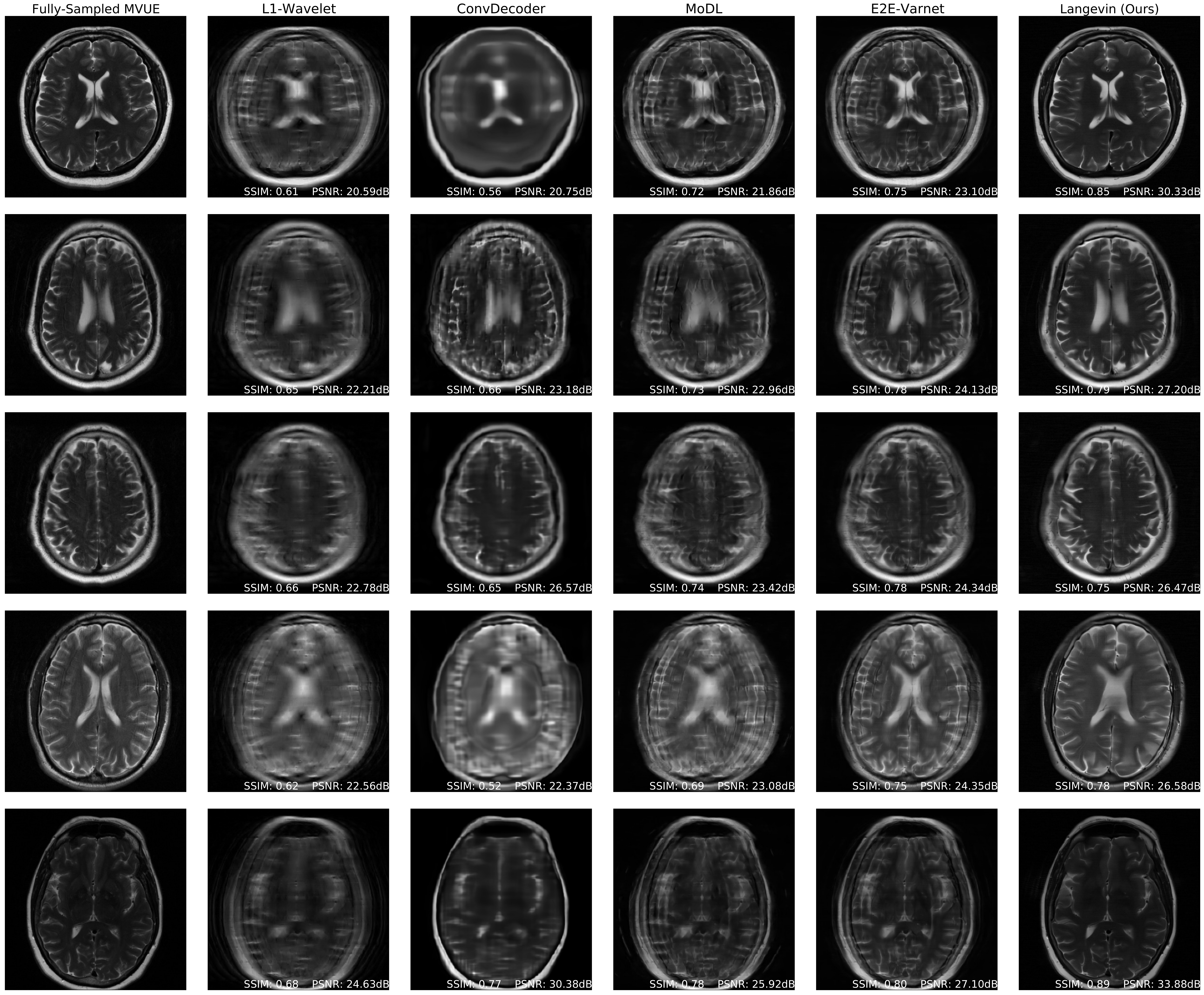}
    \caption{Brain reconstructions, at an acceleration factor of $R=12$ and an equispaced vertical mask in k-space. Our model was trained on T2-weighted brain images from the fastMRI dataset. These results show that our method has significantly fewer artifacts than baselines.}
    \label{fig:brain-in-12}
\end{figure}

\begin{figure}
    \centering
    \includegraphics[width=\columnwidth]{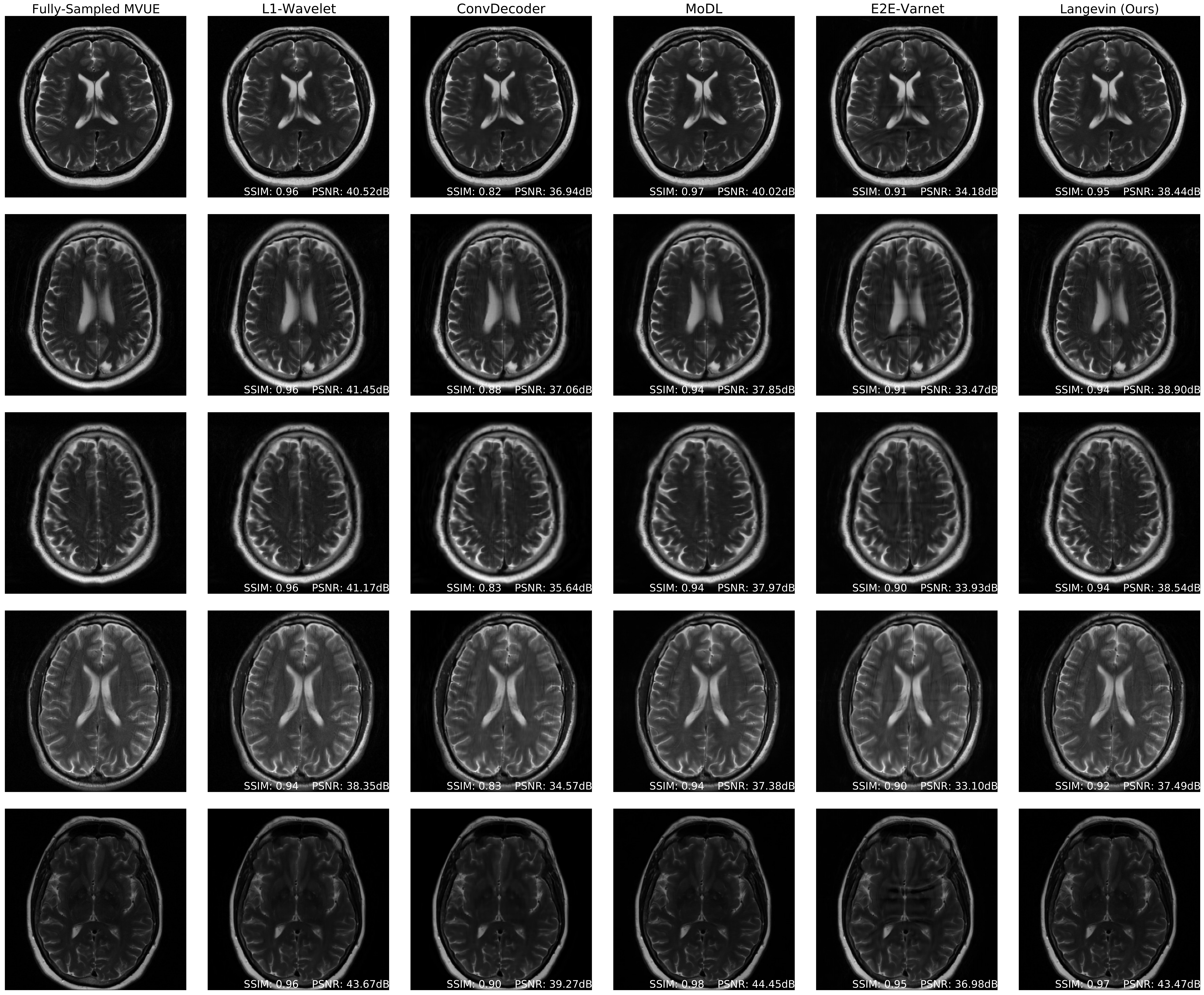}
    \caption{Brain reconstructions under a mask shift, at an acceleration of $R=3$. MoDL and E2E-VarNet were trained using an equispaced vertical mask, while these experiments were run using an equispaced \emph{horizontal} mask. Our method is robust to the mask shift, as our generative prior was trained without any knowledge of the measurement process. ConvDecoder and L1-Wavelets are untrained methods, and hence are robust to the mask shift.}
    \label{fig:brain-mask-shift-3}
\end{figure}

\clearpage

\section{Appendix: fastMRI Knee}\label{app:knees}
\begin{figure}
    \centering
    \includegraphics[width=\columnwidth]{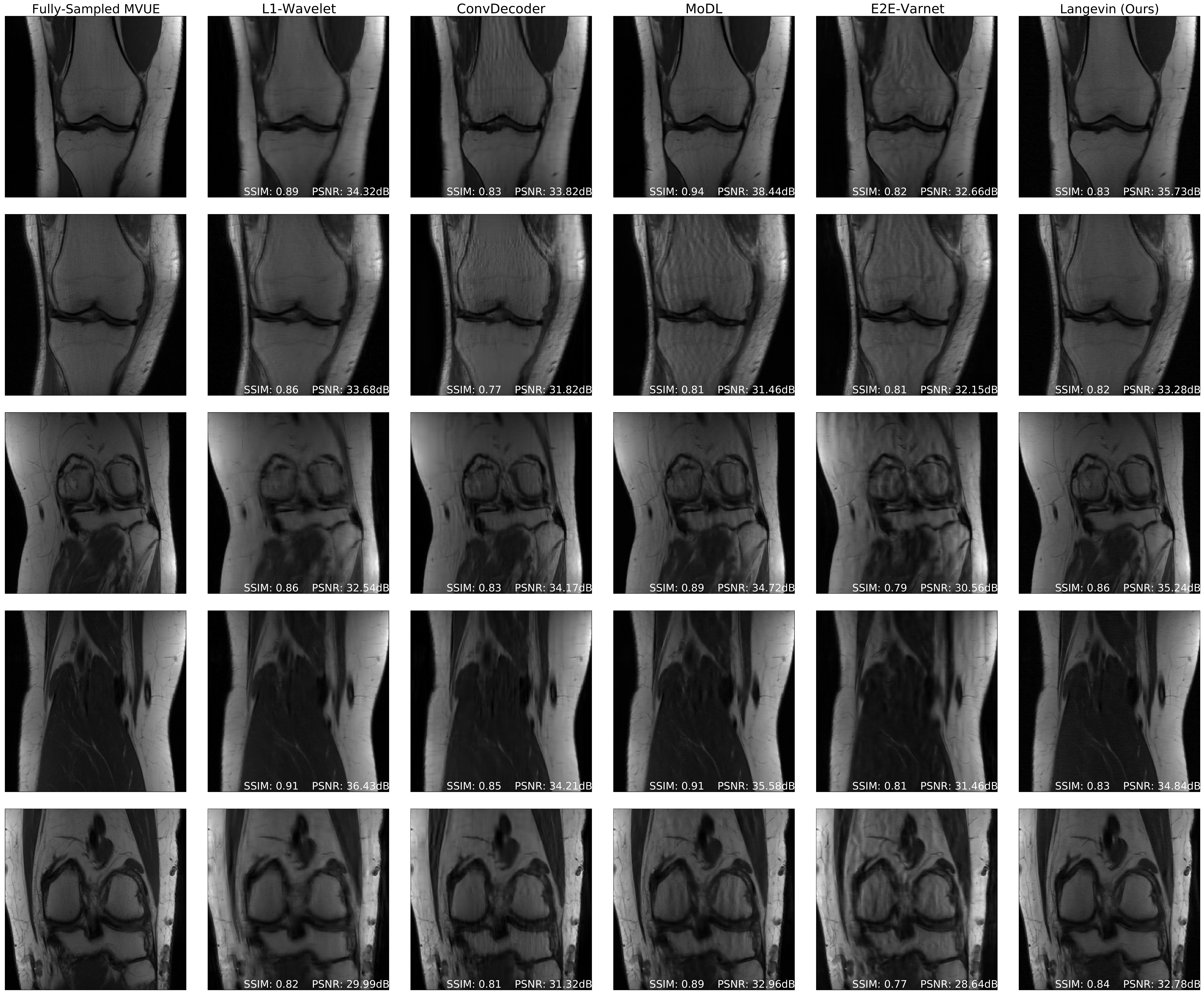}
    \caption{fastMRI knee reconstructions at an acceleration factor of $R=4$ and a random vertical mask in k-space. All methods were trained on fastMRI brains, and this shows that our method is more robust than other methods with respect to anatomy shift. 
    }
    \label{fig:knee-random-vertical-4}
\end{figure}

\begin{figure}
    \centering
    \includegraphics[width=\columnwidth]{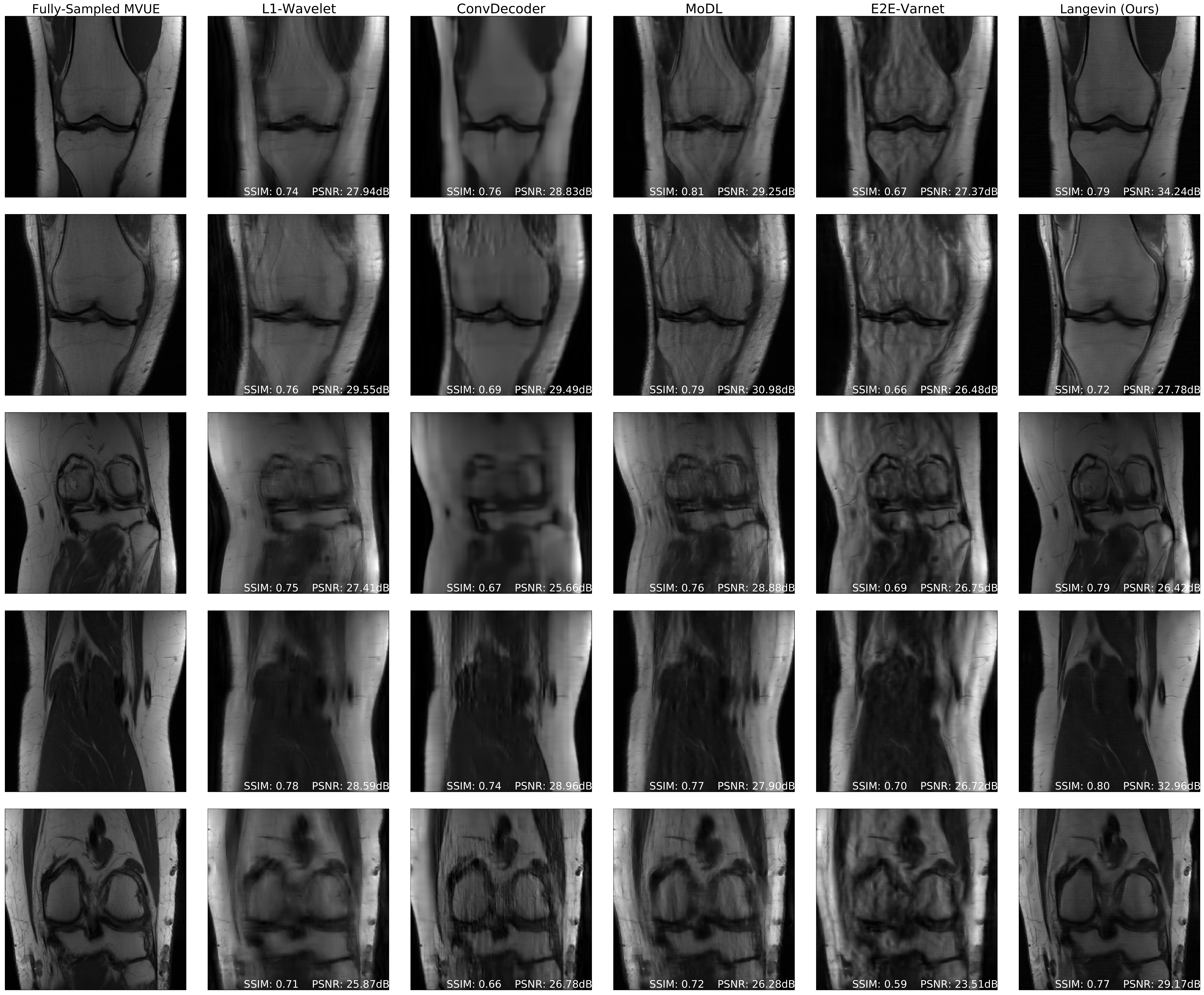}
    \caption{fastMRI knee reconstructions at an acceleration factor of $R=8$ and a random vertical mask in k-space. All methods were trained on fastMRI brains, and this shows that our method is more robust than other methods with respect to anatomy shift.
    }
    \label{fig:knee-random-vertical-8}
\end{figure}

Figure~\ref{fig:knee-random-vertical-4} and Figure~\ref{fig:knee-random-vertical-8} show further examples of proton density knee reconstructions.

\begin{figure}
    \centering
    \includegraphics[width=\columnwidth]{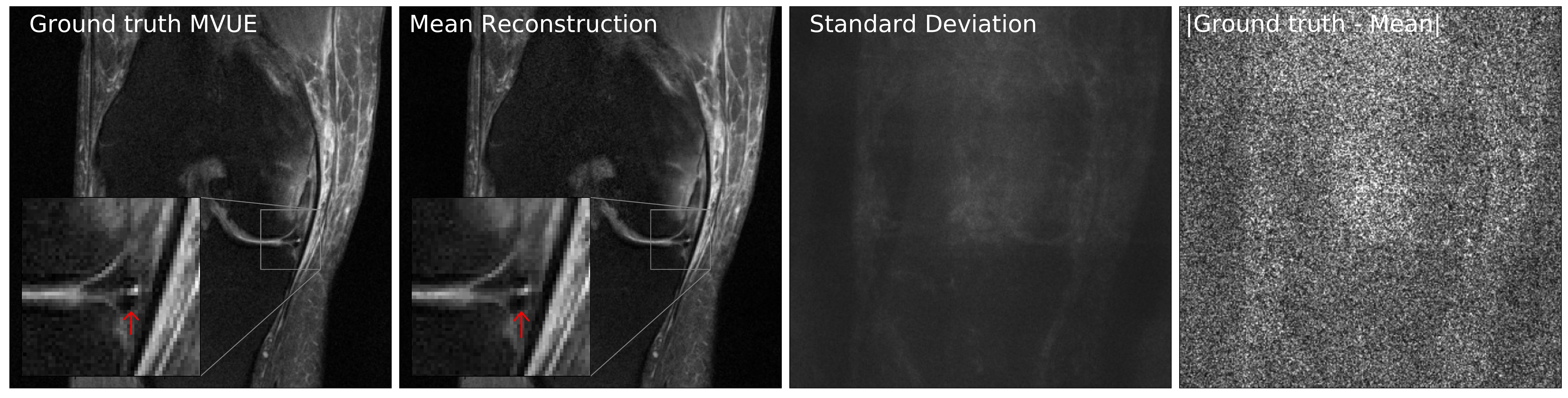}
    \caption{\small Our method successfully recovers fine details and can provide an estimate of the reconstruction error. The left column shows a knee from the fastMRI dataset, along with an annotated meniscus tear (indicated by red arrow in zoomed inset). Given measurements at an acceleration factor of $R=4$, we obtain $48$ independent reconstructions via posterior sampling. The second column shows the pixel-wise average of reconstructions, the third column shows the pixel-wise standard deviation, and the fourth column shows the magnitude of the error between the ground truth and the mean reconstruction. Note that our generative prior has never seen such pathology, as it was trained on T2-weighted brain scans.}
    \label{fig:uncertainty-app}
\end{figure}

\begin{figure}[t]
\begin{center}
  \includegraphics[width=\columnwidth]{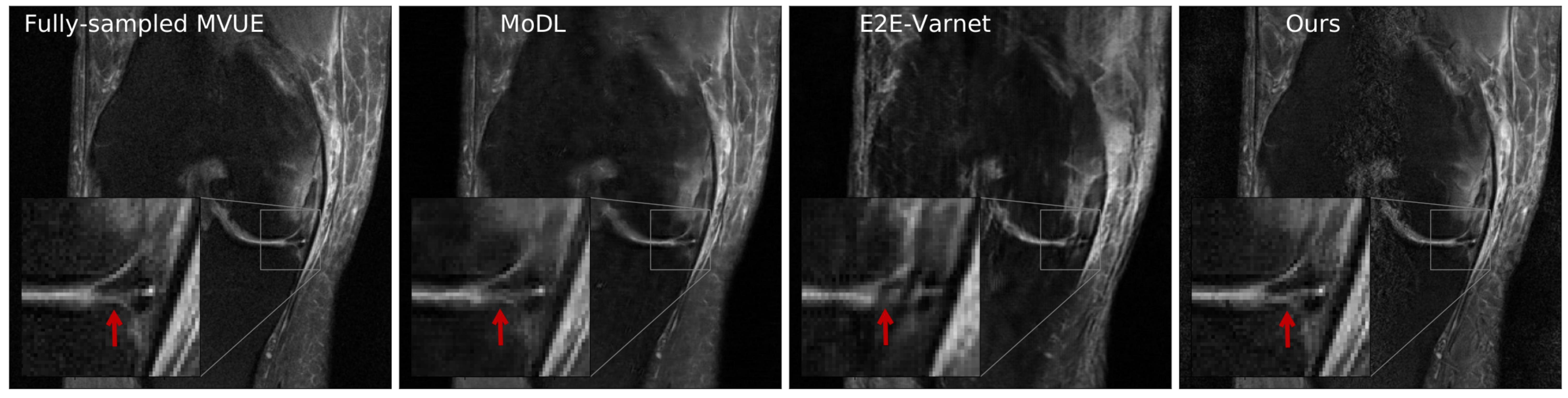}
\end{center}
\caption{\small The left column shows a knee from the fastMRI dataset, along with an annotated meniscus tear (indicated by red arrow in zoomed inset). Given measurements at an acceleration factor of $R=4$, we observe that our method preserves fine details better than the baselines. None of the methods have seen such a pathology, as they were all trained on T2-weighted brain scans.}
\label{fig:tear-baslines-1}
\end{figure}

\begin{figure}[t]
\begin{center}
  \includegraphics[width=\columnwidth]{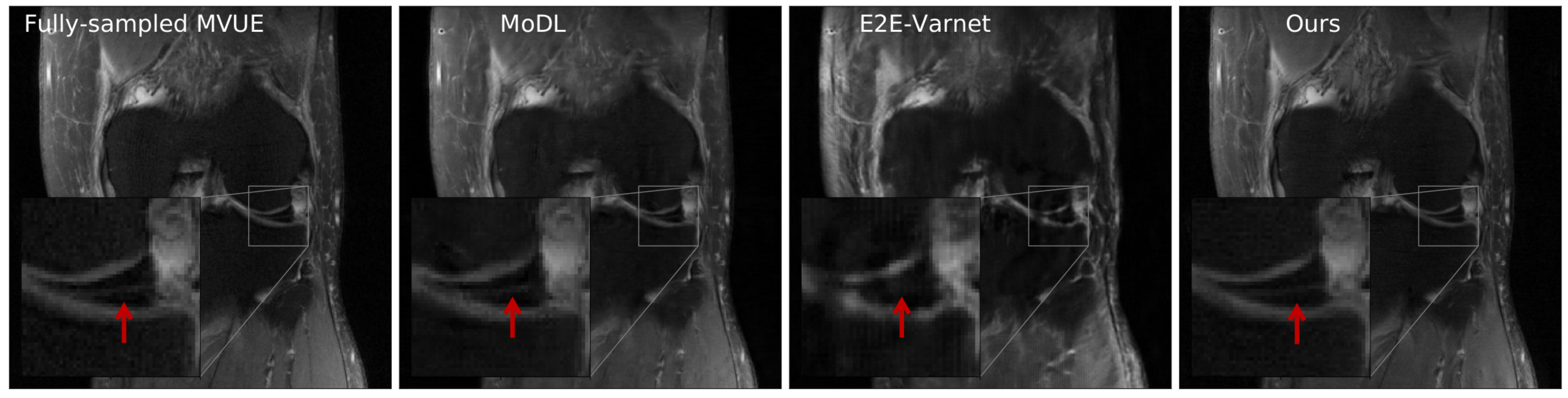}
\end{center}
\caption{\small The left column shows a knee from the fastMRI dataset, along with an annotated meniscus tear (indicated by red arrow in zoomed inset). Given measurements at an acceleration factor of $R=4$, we observe that our method preserves fine details better than the baselines. None of the methods have seen such a pathology, as they were all trained on T2-weighted brain scans.}
\label{fig:tear-baslines-3}
\end{figure}

Figure~\ref{fig:tear-baslines-1} and Figure~\ref{fig:tear-baslines-3} show comparisons of our method and baselines on knees with meniscus tears. Figure~\ref{fig:uncertainty-app} shows uncertainty estimates from our algorithm on a knee with a meniscus tear.

\begin{figure}
    \centering
    \includegraphics[width=\columnwidth]{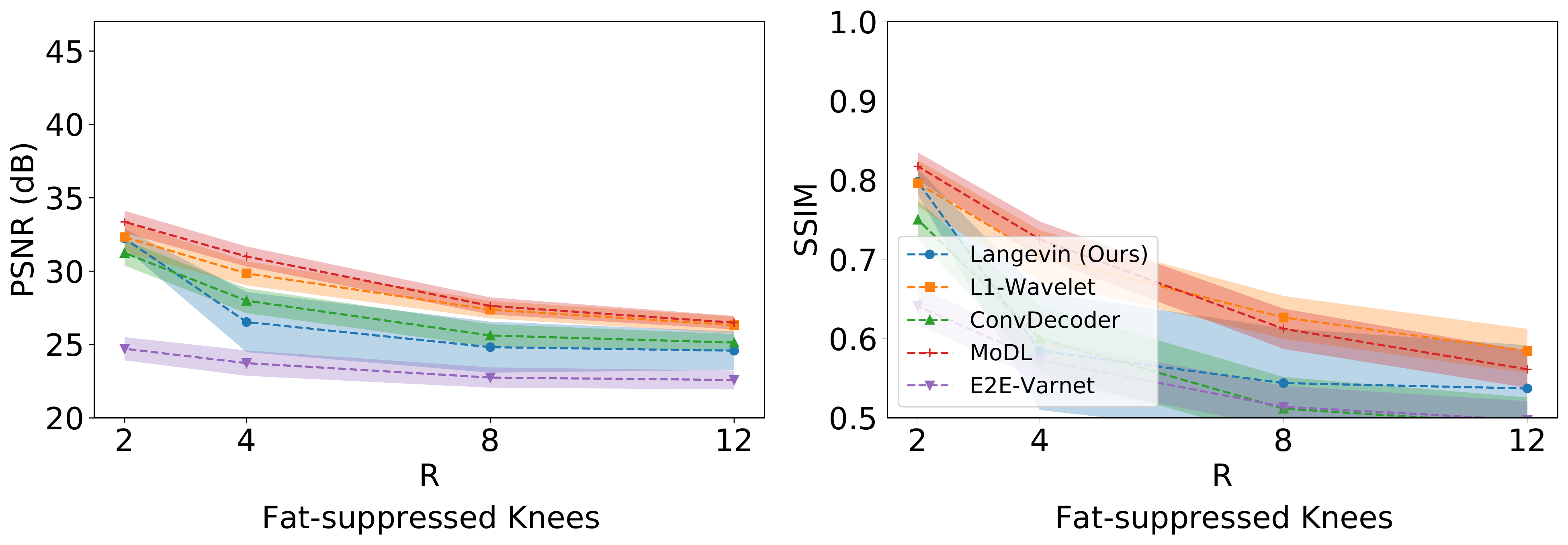}
    \caption{\small Average test PSNR and SSIM on fat-suppressed (FS) knees, across a range of acceleration factors $R$ and a random vertical mask in k-space. Higher $R$ indicates a smaller number of acquired measurements. All methods were trained on fastMRI brains. Our approach is not optimal numerically, likely due to a much lower signal-to-noise ratio in FS knees than the brain training data. However, Figures~\ref{fig:tear-baslines-1}, \ref{fig:tear-baslines-3}, \ref{fig:fs-knee-random-vertical-4}, \ref{fig:fs-knee-random-vertical-8} show that our qualitative reconstructions are competitive, and recover fine details like meniscus tears better than the deep learning baselines. Shaded regions indicate 95\% confidence intervals. Note that we trained baselines on MVUE images and hence these numerical values should not be compared with those in literature trained on RSS images (see Appendix~\ref{app:mvue-rss} for a more detailed discussion).}
    \label{fig:fs-ssim-psnr}
\end{figure}

\begin{figure}
    \centering
    \includegraphics[width=\columnwidth]{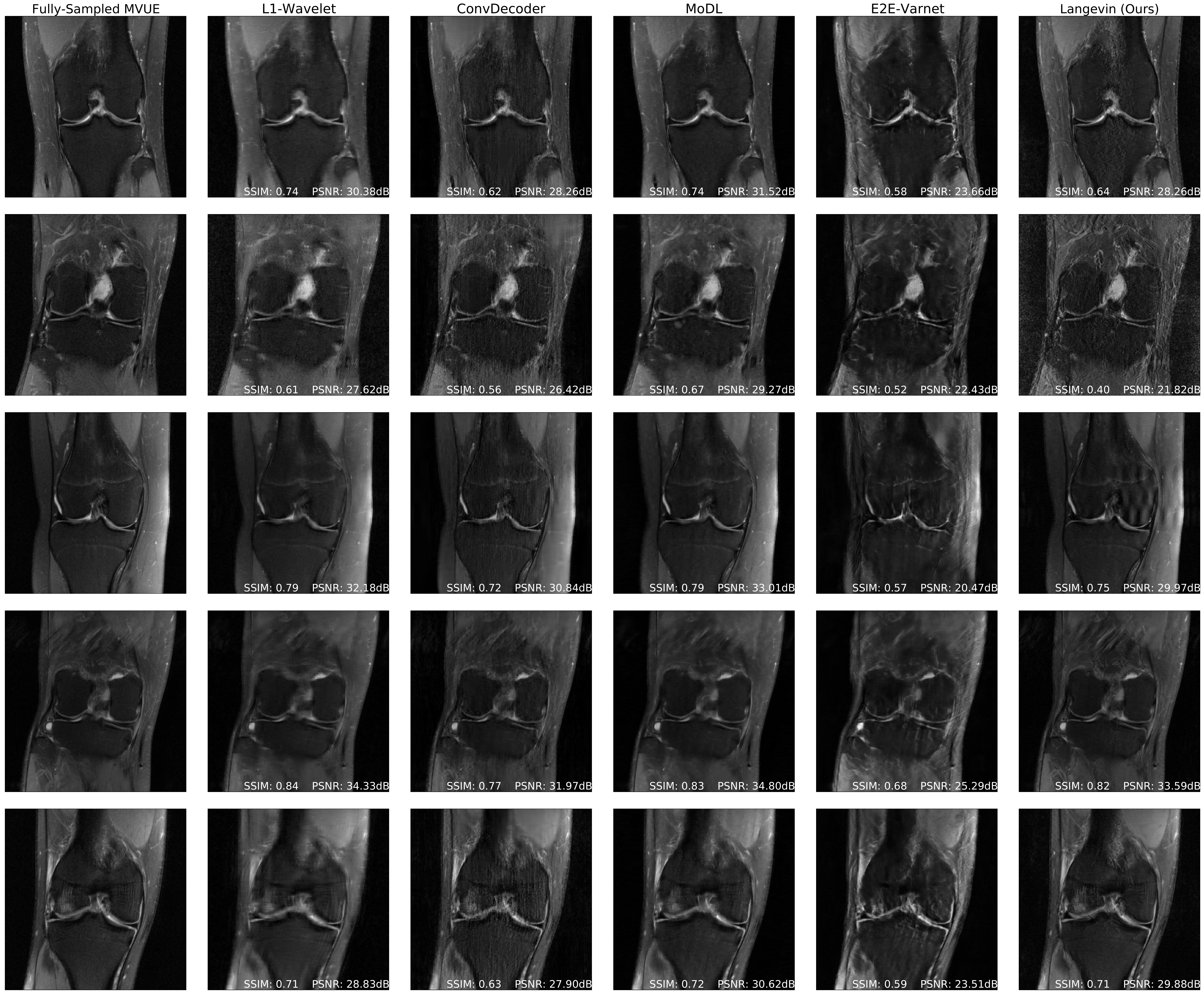}
    \caption{\small fastMRI fat-suppressed(FS) knee reconstructions at an acceleration factor of $R=4$ and a random vertical mask in k-space. All methods were trained on fastMRI brains. Our approach is not optimal numerically, likely due to a much lower signal-to-noise ratio in FS knees than the brain training data. However, the reconstructions in this figure and Figures~\ref{fig:tear-baslines-1}, \ref{fig:tear-baslines-3}, \ref{fig:fs-knee-random-vertical-8} show that our qualitative reconstructions are competitive, and recovers fine details like meniscus tears better than the deep learning baselines.}
    \label{fig:fs-knee-random-vertical-4}
\end{figure}

\begin{figure}
    \centering
    \includegraphics[width=\columnwidth]{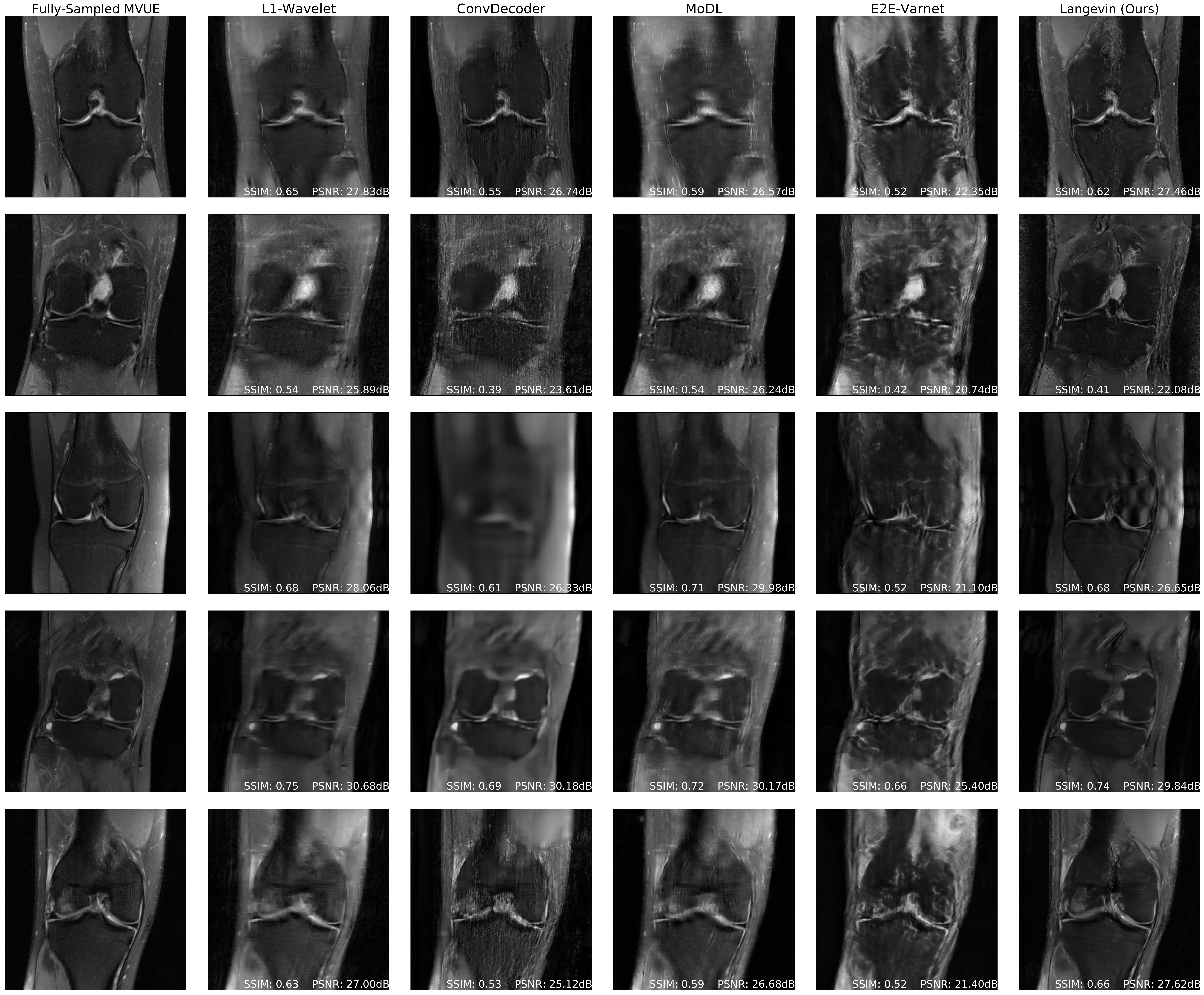}
    \caption{\small fastMRI fat-suppressed knee reconstructions at an acceleration factor of $R=8$ and a random vertical mask in k-space. All methods were trained on fastMRI brains. Our approach is not optimal numerically, likely due to a much lower signal-to-noise ratio in FS knees than the brain training data. However, the reconstructions in this figure and Figures~\ref{fig:tear-baslines-1}, \ref{fig:tear-baslines-3}, \ref{fig:fs-knee-random-vertical-4} show that our qualitative reconstructions are competitive, and recovers fine details like meniscus tears better than the deep learning baselines.}
    \label{fig:fs-knee-random-vertical-8}
\end{figure}

Figure~\ref{fig:fs-ssim-psnr} shows PSNR and SSIM on fat-suppressed(FS) knees. Our approach is not optimal numerically, likely due to a much lower signal-to-noise ratio in FS knees than the brain training data. However, Figures~\ref{fig:tear-baslines-1}, \ref{fig:tear-baslines-3}, \ref{fig:fs-knee-random-vertical-4}, \ref{fig:fs-knee-random-vertical-8} show that our qualitative reconstructions are competitive, and recovers fine details (like meniscus tears) better than the deep learning baselines.

\clearpage

\section{Appendix: Abdomen}\label{app:abdomens}
\begin{figure}
    \centering
    \includegraphics[width=\columnwidth]{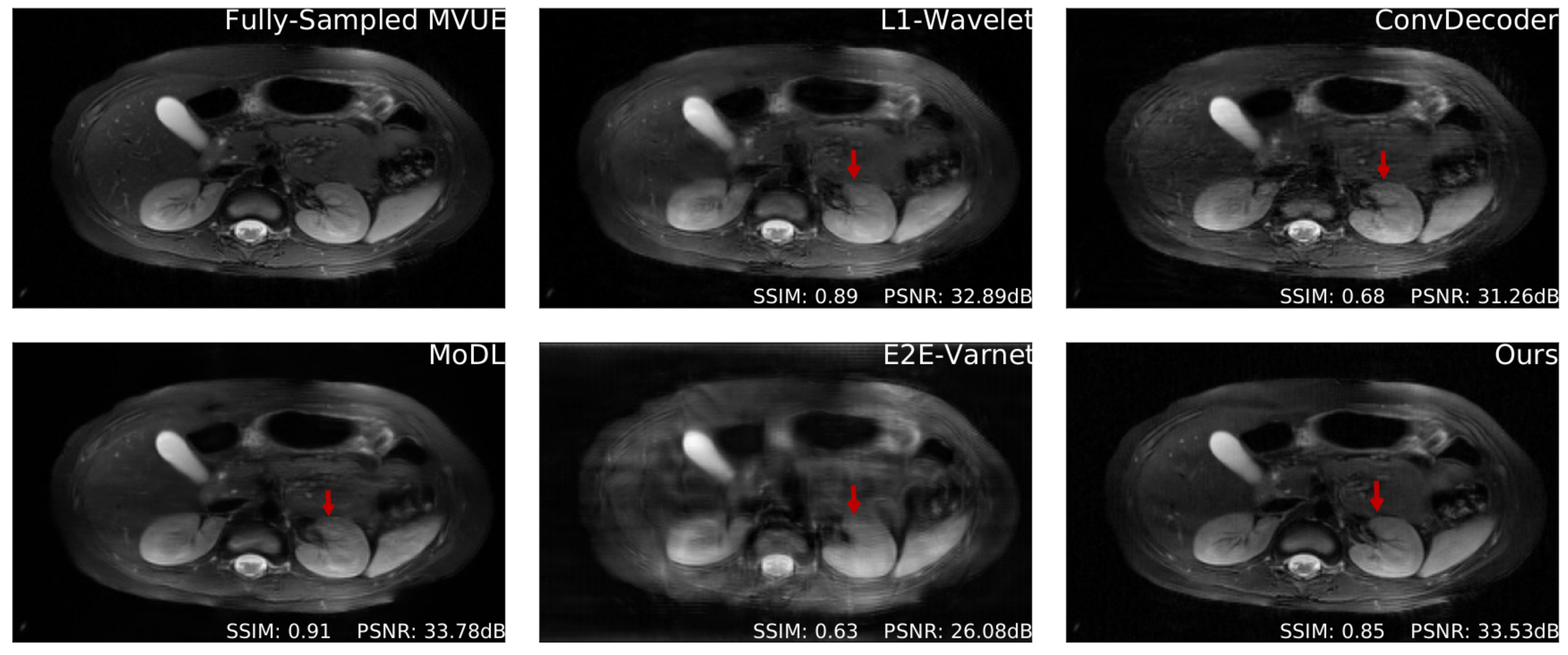}
    \caption{\small Comparative reconstructions of a 2D abdominal scan with uniform random under-sampling in the horizontal direction at $R=4$. None of the methods were trained to reconstruct abdomen MRI. Our method uses a score-based generative model trained on brain images (as explained) and obtains good reconstructions. The red arrows indicate missing details or artifacts in the kidney structure. 
    }
    \label{fig:abdomen-slice-8}
\end{figure}

Figure~\ref{fig:abdomen-slice-8} shows an additional example of a reconstructed abdominal scan. This is obtained from the same volume as the figure in the main text, and has a resolution of $158 \times 320$ voxels, but a much larger field of view, leading to a resolution shift for all models.

\section{Appendix: Stanford Knee}\label{app:3d-knees}
\begin{figure}
    \centering
    \begin{subfigure}{0.48\columnwidth}
    \centering
    \includegraphics[width=\columnwidth]{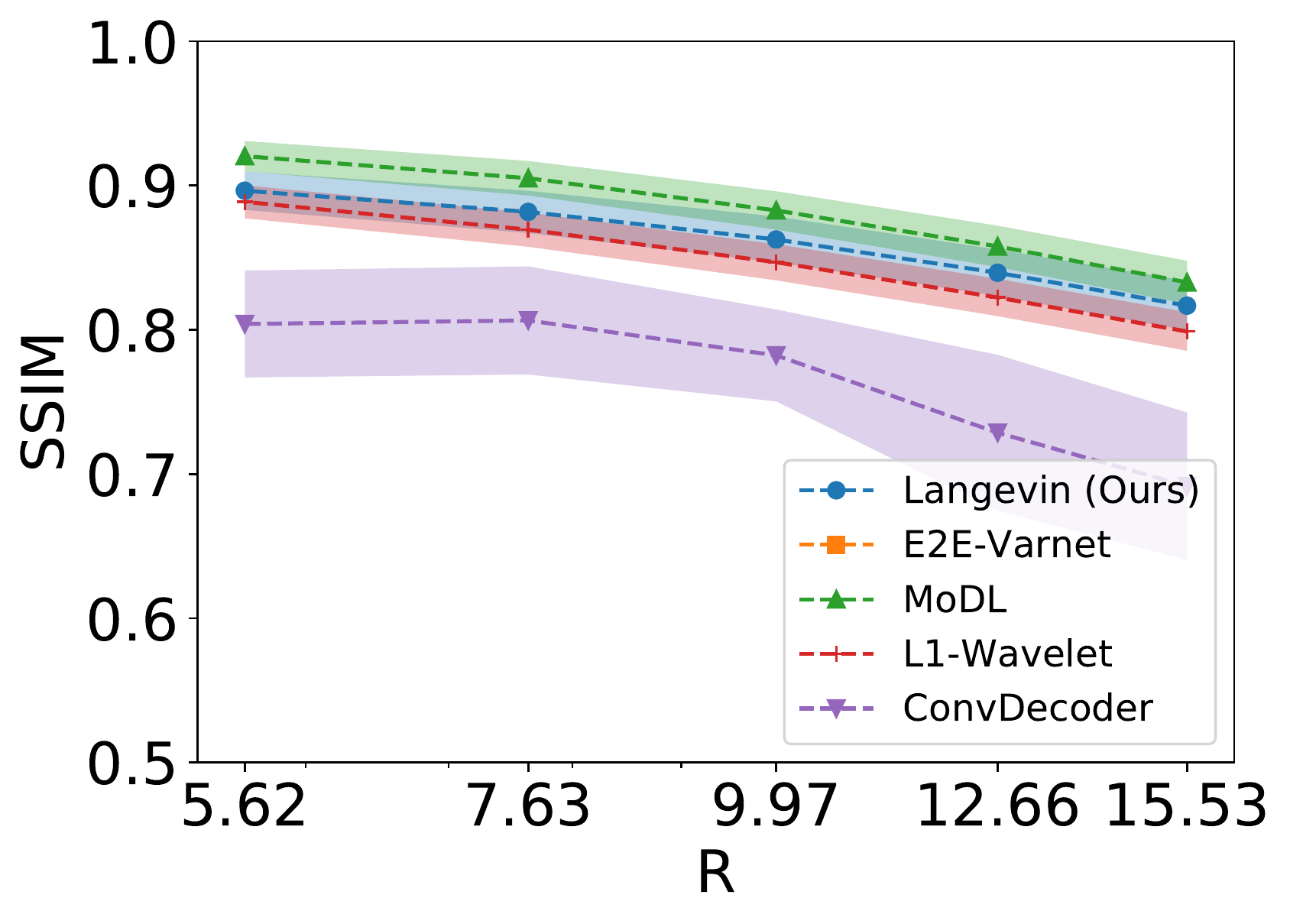}
    \caption{SSIM}
    \label{fig:stanford-ssim}
    \end{subfigure}
    \hfill
    \begin{subfigure}{0.48\columnwidth}
    \centering
    \includegraphics[width=\columnwidth]{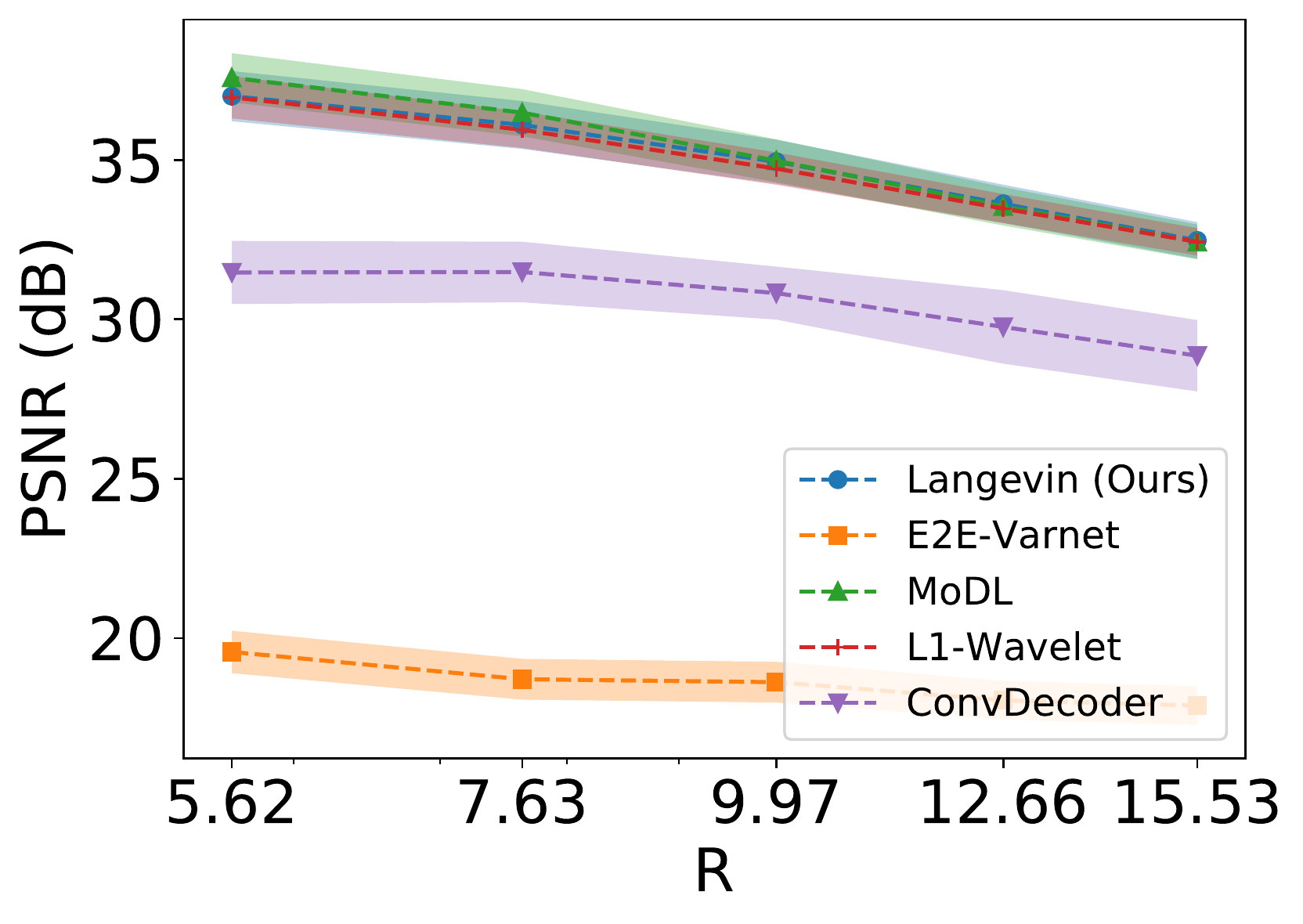}
    \caption{PSNR}
    \label{fig:stanford-psnr}
    \end{subfigure}
    \caption{Reconstruction SSIM and PSNR on Stanford Knees as a function of the acceleration $R$. This dataset is considerably different from the others, as they are 3D scans. We sample k-space measurements according to Poisson masks, which gives improved incoherence, and hence we find no statistical difference between L1-Wavelet, MoDL, and our method. Note that all hyper-parameter selection and model training was done on brains from the fastMRI dataset.}
    \label{fig:stanford-psnr-ssim}
\end{figure}

\begin{figure}
    \centering
    \includegraphics[width=\columnwidth]{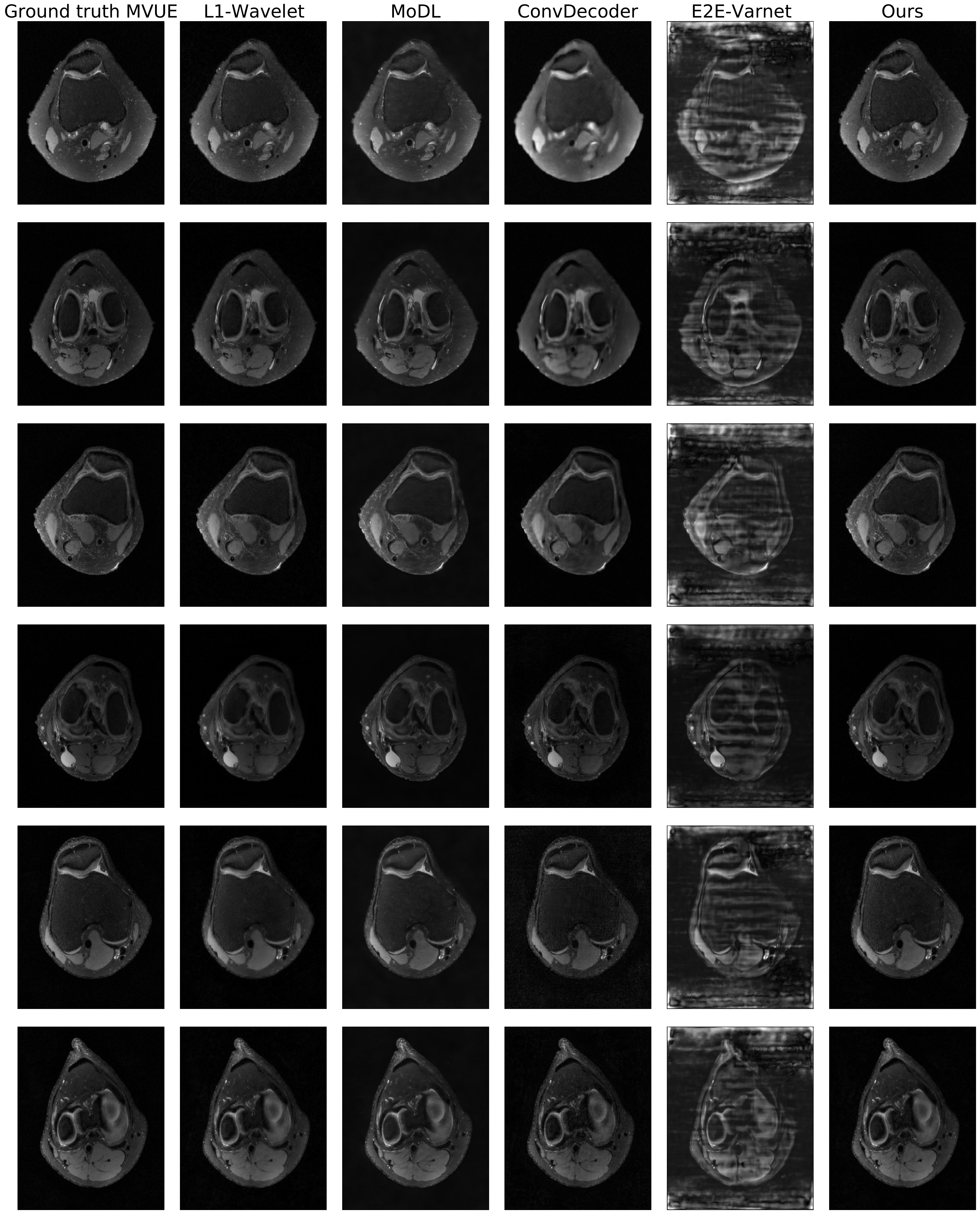}
    \caption{Qualitative reconstructions obtained by all methods on the Stanford Knees dataset at an acceleration of $R=5.62$. This dataset is considerably different from the others, as they are 3D scans. We sample k-space measurements according to Poisson masks, which gives improved incoherence, and hence we find no statistical difference between L1-Wavelet, MoDL, and our method. Note that all hyper-parameter selection and model training was done on brains from the fastMRI dataset.}
    \label{fig:stanford-recons-5.62}
\end{figure}

Figures~\ref{fig:stanford-psnr-ssim} and
\ref{fig:stanford-recons-5.62} show quantitative and qualitative
reconstruction under an anatomy shift induced by testing axial knee
scans. In this case, we first obtain a complete three-dimensional fast
spin echo (3D-FSE) knee scan from the publicly available repository at
\texttt{mridata.org}. To obtain two-dimensional slices, we apply an
IFFT operator on the readout axis and select $24$ equally spaced
slices for evaluation. Each slice has a resolution of $320 \times 256$
pixels.

\section{Appendix: Implementation}\label{app:implementation}

\subsection{Score-Based Generative Model}

\paragraph{Training the model} We use the implementation from
\url{https://github.com/ermongroup/ncsnv2}. As raw MRI scans are complex
valued, we changed the generator such that the output and input have
two channels, one each for the real and imaginary components. We did
not change the architecture otherwise.

We used the FlickrFaces (FFHQ) configs file from the NCSNv2 repo,
except we set \texttt{sigma\_begin} = 232, and \texttt{sigma\_end} =
0.0066. This is because of the smaller number of channels in MRI when
compared to FFHQ.

\paragraph{Dynamic range of the data.} MRI data exhibits a lot of
variation in the dynamic range. For example, the fastMRI dataset has
max pixel value on the order of $10^{-4}$, while the abdomen and
Stanford knee data has max pixels on the order of $10^5$. In order to
deal with this variation, during \emph{training}, we normalize each
image by the 99 percentile pixel value. During inference time, when we
do not have access to the ground-truth image, we normalize the
reconstruction using the 99 percentile pixel value of the
\emph{pseudo-inverse} complex image. We observe that this heuristic is sufficient to
get good results.

\paragraph{Invariance to image shapes.} Due to the convolutional
nature of NCSNv2, although we trained on $384\times 384$ images, we
can still apply them to knees, T1-weighted \& FLAIR brains, and
abdomens, although all of these have different dimension shapes. 

\paragraph{Hyperparameters} We tuned our hyperparameters on two validation brain scans, at an acceleration of $R=4$. We then reused these hyperparameters on \emph{all anatomies, all accelerations}. Please see our GitHub link: 
\url{https://github.com/utcsilab/csgm-mri-langevin} for the hyperparameter values.

\subsection{E2E-VarNet Baseline}
We use the architecture publicly available in the fastMRI official repository. The backbone for the image reconstruction network is a U-Net with a depth of four stages, and $18$ hidden channels in the first stage, for a total of $29$ million learnable parameters. This model also include a smaller deep neural network that is used to estimate the sensitivity maps. This is also a U-Net, with four stages, but only eight hidden channels after the first stage, for an additional $0.7$ million parameters. The model is trained for a number of $12$ unrolls, and separate image networks are used at each unroll.

We train this model from scratch for a number of $40$ epochs, using an Adam optimizer with default PyTorch parameters and a learning rate of $2\mathrm{e}{-4}$, decayed by $0.5$ after $20$ epochs, as well as gradient clipping to a maximum magnitude of $1$. We use the fully-sampled MVUE reconstructions from the brain T2 contrast in fastMRI to train all methods. We use a batch size of $1$ and a supervised SSIM loss between the absolute values of ground truth MVUE and the absolute value of the complex output of the network at acceleration factors $R=\{3,6\}$ (chosen with equal probability), using a vertical, equispaced sampling pattern, same as all other baselines.

Finally, it is worth mentioning that the network used to estimate the sensitivity maps explicitly uses the fully-sampled, vertical ACS region, as shown in Figure~\ref{fig:example_masks}, both during training and inference. This makes testing with other mask patterns non-trivial for this baseline. To alleviate this, we always feed the image obtained from the \textit{vertical} ACS region (for example, in the case of horizontal masks, we intentionally zero out other sampled lines that would fall in this region), to not introduce incoherent aliasing in this image.

\subsection{MoDL Baseline}
We use the PyTorch MoDL implementation publicly available at \url{https://github.com/utcsilab/deep-jsense} and train a MoDL model that uses a backbone residual network with a depth of six layers, three equispaced residual connections (that feed hidden signals from the first three layers to the last three layers) and $64$ hidden channels, with a total of $220000$ trainable parameters. Unlike E2E-VarNet, the same backbone network is used across all unrolls, and the data consistency term is given by a Conjugate Gradient (CG) operator, truncated to six steps.

We train MoDL for a number of six unrolls, leading to a total of $36$ CG steps and six network applications in the unroll. We use the Adam optimizer with default PyTorch parameters and learning rate $2\mathrm{e}{-4}$, as well as gradient clipping to a maximum magnitude of $1$. We train for $15$ epochs and decay the learning rate by $0.5$ after $8$ epochs, using a batch size of $1$ on exactly the same T2 brain scans as all methods and a supervised SSIM loss at $R=\{3,6\}$ (chosen with equal probability) between the magnitude of the ground-truth MVUE image and the magnitude of the complex network output. We find that, although relatively small, the backbone network architecture is sufficient to achieve good in-distribution reconstruction, and serve as a strong baseline.

Since MoDL and all other methods (including ours) except E2E-VarNet, require external sensitivity map estimates to be provided to them, we use the ESPIRiT algorithm from the BART toolbox \cite{bart} without any eigenvalue cropping to estimate a single set of sensitivity maps, one for each coil.

\section{Appendix: Radiologist Study}\label{app:radiologist}
We performed a preliminary image quality assessment experiment with two board-certified radiologists and a faculty member that uses neuro-imaging in their research.

The three external experts were not involved with our research and have performed the image quality assessment blindly. Each of them was presented with ten scans from the following anatomies and scan parameters: abdominal scans, knee scans and brain scans with a horizontal readout direction, leading to a total of 30 quality assessment questions. Note that all anatomies represent test-time distributional shifts in at least one aspect.

In each question, the experts were shown four images:
\begin{itemize}
    \item The fully-sampled reference image, explicitly marked as "Reference".
    \item The results of three reconstruction algorithms at acceleration factor R=3: MoDL, ConvDecoder and our method. The order of the reconstructions was shuffled for each question, and the reconstructions were labeled as "1", "2" and "3".
\end{itemize}

We chose to compare with MoDL and ConvDecoder since these method had the best overall quantitative and qualitative (according to our own pre-assessment) robust performance. The participants were instructed to rank the three reconstructions from best to worst quality, while using the "Reference" image as a perceptual guideline. Table~\ref{tab:radiologist-scores} shows the average and standard deviation (in parentheses) of the ranking for each anatomy, obtained using a total of 30 data points (3 participants x 10 scans per anatomy).
\begin{table}
    \centering
    \begin{tabular}{|c|c|c|c|}
        \hline
        Anatomy & MoDL & ConvDec & Ours \\
        \hline
        Knee	& $1.87 (0.34)$ & $2.97 (0.18)$	& $1.17 (0.45)$ \\
        Abdomen & $1.87 (0.76)$ & $2.17 (0.93)$ & $1.97 (0.71)$ \\
        Brain	& $2.00 (0.82)$ & $2.07 (0.77)$ & $1.93 (0.85)$ \\
        \hline
    \end{tabular}
    \caption{Ranking of algorithms by experts. A lower ranking is better: the best possible ranking is 1, and the worst 3. The values show the average and standard deviation (in parentheses) of the ranking for each anatomy, using a total of 30 data points (3 participants x 10 scans per anatomy).}
    \label{tab:radiologist-scores}
\end{table}

In Table~\ref{tab:radiologist-scores}, a lower ranking is better, the best possible ranking is 1, and the worst 3. We draw the following conclusions:
\begin{itemize}
    \item Participants consistently ranked our method as best on the knee scans, which supports the distributional shift robustness claimed in the main paper, and detailed in Appendices~\ref{app:knees},~\ref{app:abdomens} and~\ref{app:metrics}.
    \item Participants did not perceive a significant difference between all methods when applied to abdominal or brain scans with a horizontal phase encode direction. In the brain case, this supports the qualitative results shown in Appendix~\ref{app:brains}, Figure~\ref{fig:brain-in-3}.
    \item In the abdominal case, this partially correlates with Figure~\ref{fig:main-psnr}c, regarding the quantitative tie between our approach and MoDL.
\end{itemize}

To quantify the statistical significance of the above results, we perform a Wilcoxson Rank Sum test to determine if the rankings of different algorithms are drawn from different populations. We evaluate if our proposed method leads to different rankings than MoDL and the ConvDecoder, and show the p-values in Table~\ref{tab:radiologist-pval}.
\begin{table}
    \centering
    \begin{tabular}{|c|c|c|}
    \hline
    Anatomy & Ours vs. MoDL	& Ours vs. ConvDec \\
    \hline
    Knee	& $1.53e-10$ & $2.77e-6$ \\
    Abdomen	& $0.610$	& $0.340$ \\
    Brain	& $0.767$	& $0.550$ \\
    \hline
    \end{tabular}
    \caption{p-values from the Wilcoxson Rank Sum test to determine if the rankings of different algorithms are drawn from different populations. There is a significant difference in the case of knees, and no significant difference in the case of abdomens and brains.}
    \label{tab:radiologist-pval}
\end{table}

The results show a significant difference in the case of knees, while no significant difference is present for abdomen and brain. Finally, to evaluate inter-observer agreement between the three reviewers, we calculated the intra-class correlation (ICC) coefficient separately for each anatomy by aggregating the ten questions related to that anatomy and evaluating the ICC2 coefficient~\cite{pingouin-stats} in a pairwise manner at a $5\%$ significance level.

The results are shown in Table~\ref{tab:radiologist-icc2}, where we also include the p-value and the $95\%$ confidence interval for the ICC2 estimate. This indicates that there exists a very strong consensus regarding the ranking on the knee anatomy, while for abdomen and brain this consensus is much weaker, which together with Table~\ref{tab:radiologist-pval} indicates that the images were considered equivalent.
\begin{table}
    \centering
    \begin{tabular}{|c|c|c|c|}
    \hline
    Anatomy & ICC2	& p-value & 95\% CI \\
    \hline
    Knee	& $0.980$  & $0.0004$ & $[0.81, 1]$ \\
    Abdomen	& $-0.222$ & $0.576$  & $[-0.89, 0.92]$ \\
    Brain	& $-0.818$ & $0.907$  & $[-0.98, 0.59]$ \\
    \hline
    \end{tabular}
    \caption{p-values and confidence intervals for differences in ranking between our method and baselines.}
    \label{tab:radiologist-icc2}
\end{table}

This preliminary image quality assessment gives additional evidence
(in addition to the quantitative metrics of SSIM and PSNR) that our
method maintains robustness to distribution shifts at test time. As
our quantitative results show, other methods maintain robustness in
some but not all cases. Due to time limitations, we were not able to
ask the reviewers to evaluate every algorithm and every distribution
shift including different levels of acceleration. We stress that this
preliminary study is not a substitute for a rigorous clinical
evaluation which is necessary before considering using our proposed
method in a clinical setting.

\end{document}